\documentclass[conference]{IEEEtran}
\usepackage{times}

\usepackage[numbers,sort]{natbib}
\usepackage[bookmarks=true]{hyperref}
\newenvironment{tightlist}%
{\begin{list}{$\bullet$}{%
    \setlength{\topsep}{0in}
    \setlength{\partopsep}{0in}
    \setlength{\itemsep}{0in}
    \setlength{\parsep}{0in}
    \setlength{\leftmargin}{1.5em}
    \setlength{\rightmargin}{0in}
}
}%
{\end{list}
}

\usepackage[ruled,vlined]{algorithm2e}

\usepackage{amsmath}
\usepackage{booktabs}
\usepackage{graphicx}
\usepackage{wrapfig}
\usepackage{comment}
\usepackage{float}
\usepackage{amsthm}
\usepackage{amssymb}
\usepackage{mathtools}
\usepackage{float}
\usepackage[normalem]{ulem}
\usepackage{wrapfig}
\usepackage{listings}
\usepackage{color}
\usepackage{multicol}
\usepackage{footnote}
\usepackage[dvipsnames]{xcolor}

\usepackage[small]{caption}

\newtheorem{assumption}{Assumption}%[theorem]

\theoremstyle{definition}
\newtheorem{definition}{Definition}

\newcommand{\states}{\mathcal{X}}
\newcommand{\state}{x}
\newcommand{\staterv}{X}
\newcommand{\parameterizedskills}{\mathcal{U}}
\newcommand{\parameterizedskill}{u}
\newcommand{\actions}{\mathcal{A}}
\newcommand{\action}{a}
\newcommand{\actionrv}{A}
\newcommand{\goal}{g}
\newcommand{\goalrv}{G}
\newcommand{\horizoneval}{H_{\text{eval}}}

\newcommand{\parameterprior}{\pi^0}
\newcommand{\parameterpolicy}{\pi}
\newcommand{\parameterpolicies}{\Pi}

\newcommand{\exploreparameterpolicy}{\pi^+}
\newcommand{\competencerv}{C}
\newcommand{\outcome}{S}

\DeclareMathOperator*{\argmin}{argmin}
\DeclareMathOperator*{\argmax}{argmax}
\DeclarePairedDelimiterX{\infdivx}[2]{(}{)}{%
  #1\;\delimsize\|\;#2%
}

\newcommand{\abstractfn}{\texttt{abstract}}
\newcommand{\termination}{\beta}
\newcommand{\successfn}{J}
\newcommand{\precondition}{I}
\newcommand{\abstractstate}{s}

\newcommand{\dataset}{\mathcal{D}}
\newcommand{\energyfn}{E}
\newcommand{\lowlevelpolicy}{\mu}

\newcommand{\competence}{c}

\makeatletter
\apptocmd\@maketitle{{\myfigure{}\par}}{}{}
\newcommand{\removelatexerror}{\let\@latex@error\@gobble}
\makeatother

\newcommand\blfootnote[1]{%
  \begingroup
  \renewcommand\thefootnote{}\footnote{#1}%
  \addtocounter{footnote}{-1}%
  \endgroup
}

\usepackage{fancyvrb}

\begin{document}
\newcommand\myfigure{%
\centering
\noindent
\vspace{-0.5em}
\includegraphics[width=\textwidth]{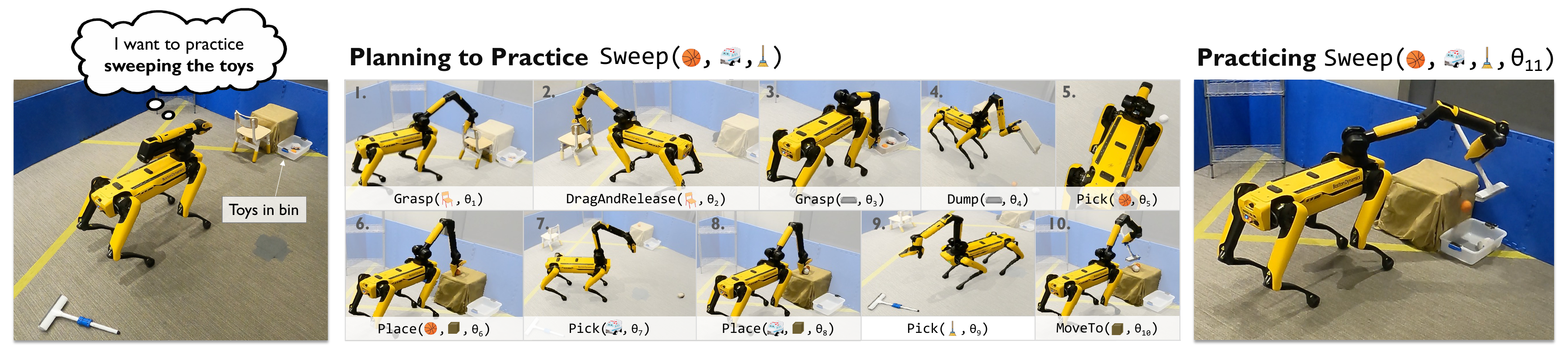}
\vspace{-1.25em}
\captionof{figure}{\textbf{Planning to practice.} To practice a skill, the robot needs to be in a state where the skill can be initiated. Here, the robot plans to practice sweeping two toys into a bin from its initial state (left) in the Cleanup Playroom environment. This requires chaining up to 19 skills (middle), some of which are omitted for brevity, before practicing (right). Some skill object parameters are also omitted.}
\vspace{-0.5em}
\label{fig:teaser}
% NOTE: I have no idea why this is necessary, but without it, the figure number skips...
\setcounter{figure}{1}
}

\title{Practice Makes Perfect: Planning to \\ Learn Skill Parameter Policies}

% You will get a Paper-ID when submitting a pdf file to the conference system
% \author{Author Names Omitted for Anonymous Review. Paper-ID 90}

% \author{\authorblockN{Nishanth Kumar \authorrefmark{2}\authorrefmark{3}} \thanks{Equal contribution.}
% \authorblockA{\texttt{njk@csail.mit.edu}}
% \and
% \authorblockN{Tom Silver \authorrefmark{2}\authorrefmark{3}}
% \authorblockA{\texttt{tslvr@csail.mit.edu}}
% \and
% \authorblockN{Willie McClinton \authorrefmark{2}\authorrefmark{3}}
% \authorblockA{\texttt{wmcclinton@csail.mit.edu}}
% \and
% \authorblockN{Linfeng Zhao \authorrefmark{2}\authorrefmark{4}}
% \authorblockA{\texttt{zhao.linf@northeastern.edu}}
% \and
% \authorblockN{Tom\'{a}s Lozano-P\'{e}rez \authorrefmark{3}}
% \authorblockA{\texttt{tlp@csail.mit.edu}}
% \and
% \authorblockN{Leslie Kaelbling \authorrefmark{3}}
% \authorblockA{\texttt{lpk@csail.mit.edu}}
% \and
% \authorblockN{Jennifer Barry \authorrefmark{2}}
% \authorblockA{\texttt{jbarry@theaiinstitute.com}}
% }

% avoiding spaces at the end of the author lines is not a problem with
% conference papers because we don't use \thanks or \IEEEmembership

\author{\authorblockN{Nishanth Kumar\authorrefmark{1}\authorrefmark{2}\authorrefmark{3},
Tom Silver\authorrefmark{1}\authorrefmark{2}\authorrefmark{3},
Willie McClinton\authorrefmark{2}\authorrefmark{3}, 
Linfeng Zhao\authorrefmark{2}\authorrefmark{4}, \\
Stephen Proulx\authorrefmark{2},
Tom\'{a}s Lozano-P\'{e}rez\authorrefmark{3},
Leslie Pack Kaelbling\authorrefmark{3} and
Jennifer Barry\authorrefmark{2}}
\authorblockA{\authorrefmark{2}The AI Institute, \authorrefmark{3}MIT CSAIL, \authorrefmark{4}Northeastern University}}

\maketitle

\begin{abstract}
One promising approach towards effective robot decision making in complex, long-horizon tasks is to sequence together \emph{parameterized skills}.
We consider a setting where a robot is initially equipped with (1) a library of parameterized skills, (2) an AI planner for sequencing together the skills given a goal, and (3) a very general prior distribution for selecting skill parameters.
Once deployed, the robot should rapidly and autonomously learn to improve its performance by specializing its skill parameter selection policy to the particular objects, goals, and constraints in its environment.
In this work, we focus on the active learning problem of choosing which skills to \emph{practice} to maximize expected future task success.
We propose that the robot should \emph{estimate} the competence of each skill, \emph{extrapolate} the competence (asking: ``how much would the competence improve through practice?''), and \emph{situate} the skill in the task distribution through competence-aware planning.
This approach is implemented within a fully autonomous system where the robot repeatedly plans, practices, and learns without any environment resets.
Through experiments in simulation, we find that our approach learns effective parameter policies more sample-efficiently than several baselines.
Experiments in the real-world demonstrate our approach's ability to handle noise from perception and control and improve the robot's ability to solve two long-horizon mobile-manipulation tasks after a few hours of autonomous practice.
Project website: \url{http://ees.csail.mit.edu}
\vspace{1em}
\blfootnote{$^*$Equal contribution. Work done during internship at The AI Institute. Correspondence to \{njk, tslvr\}@csail.mit.edu.}
\end{abstract}
\IEEEpeerreviewmaketitle

\section{Introduction}
\label{sec:intro}

Given the recent progress in robot skill learning and design \cite{chi2023diffusionpolicy,wan2023lotus,zhao2023learning,ma2023eureka,fu2024mobile}, we are quickly approaching a future where robots will arrive at their deployment sites equipped with a library of general-purpose skills.
Each robot will sequentially compose these skills in different ways to accomplish long-horizon tasks that will vary considerably between deployment sites.
As the robot gathers experience during deployment, it should get better over time.
In particular, the robot should learn to \emph{rapidly specialize} its skills to the unique objects, goals, and constraints that it repeatedly encounters during deployment.

In this work, we consider skills that are continuously parameterized and we focus on \emph{parameter policy learning}~\cite{masson2016reinforcement,ames2018learning,dalal2021accelerating,nasiriany2022maple,fang2023active} as a mechanism for rapidly specializing skills.
For example, a ``pick'' skill may be parameterized by a relative grasp and a ``sweep'' skill by a sweeping velocity (Figure~\ref{fig:teaser}).
Starting from general-purpose priors~\cite{singh2021parrot,pertsch2021accelerating,biza2021action,gupta2023bootstrapped}, we want the robot to quickly learn specialized policies for selecting grasps, push velocities, and other skill parameters.
Following previous work~\cite{konidaris2018skills,ames2018learning,silver2022learning}, we consider parameterized skills that are (extended) \emph{options}~\cite{sutton1999between}; each skill has an initiation condition, a parameterized controller, a termination condition, and a success condition.
For example, a ``place'' skill can be initiated when the robot is holding an object and facing a surface; the skill terminates after the robot opens its gripper; and the skill is successful if the object is subsequently stably resting on the surface.
Options are closely related to AI planning operators~\cite{konidaris2018skills,silver2022learning} and
we can leverage this relationship to efficiently \emph{plan} a sequence of skills to reach a goal~\cite{helmert2006fast}.

We consider parameter policy learning in the context of reset-free online learning~\cite{thrun1995lifelong,lu2020reset,gupta2021reset,mendez2023embodied} where the robot alternates between solving given tasks (\emph{task time}) and taking actions of its choosing (\emph{free time}).
For example, a given task might be to ``clear objects off the table'' (Figure~\ref{fig:teaser}).
We focus on free time and ask: how should the robot select actions so that, after learning parameter policies from the collected experience, the likelihood of solving given tasks in the future is maximized?
This is an \emph{embodied active learning} problem~\cite{daniel2014active,noseworthy2021active,li2023embodied,mendez2023embodied}, which is distinct from standard active learning~\cite{settles2011theories} in that the robot must reason sequentially.
For example, to collect one ``sweep'' data point in our experiments, the robot needs to execute up to 19 skills in sequence to reach a state where sweeping is possible (Figure~\ref{fig:teaser}).
This setting is also related to \emph{exploration} in the reinforcement learning literature~\citep{kearns2002near,pathak2019selfsupervised,colas2019curious,bougie2020skill,amin2021survey}; we consider baselines from that literature in experiments.
Compared to end-to-end RL though, our setting has significantly more structure, which we can leverage to achieve much more sample-efficient learning, especially over long-horizon tasks.

To learn parameter policies through embodied active learning, we consider \emph{planning to practice parameterized skills}.
During free time, the robot repeatedly selects a skill, plans to a state where that skill can be initiated, practices the skill (selects continuous parameters to try), records the success condition outcome, and then updates its parameter policy accordingly.
The central question is: how should the robot decide what skills to practice?
One approach would be to practice the skill with the lowest \emph{competence}~\cite{stout2010competence,da2014active}, that is, the skill that most often fails to achieve its success condition.
But that skill may be impossible to improve or irrelevant to the given tasks.

We propose that the robot should instead \emph{practice the skill whose predicted competence improvement would maximally benefit the overall task distribution}.
Implementing this skill selection strategy requires three steps: \emph{estimating} current skill competence; \emph{extrapolating} the competence by predicting how much it would hypothetically improve through practice; and \emph{situating} the competence in the task distribution by predicting how overall task success rates would hypothetically improve.
We propose a Beta-Bernoulli time series model to estimate and extrapolate skill competence and use cost-aware AI planning~\cite{helmert2006fast} to situate the competence in the task distribution.

In experiments, we evaluate the extent to which our \textbf{Estimate, Extrapolate \& Situate (EES)} approach enables the robot to make efficient use of its free time as measured by its success rate during task time.
In three simulated environments, we compare to seven baselines and find that EES is consistently the most sample-efficient.
We also implement EES in two real mobile manipulation environments using a Boston Dynamics Spot robot with an arm (Figures~\ref{fig:teaser}, \ref{fig:ball-ring-env}).
In these environments, the robot plans and practices autonomously for several hours, coping with noise inherent to real-world perception and control, and rapidly improves its ability to solve long-horizon mobile-manipulation tasks.

\section{Problem Setting}
\label{sec:problem-setting}

This paper proposes a method for active practicing in the context of a robot system that has mechanisms for planning and learning.
In this section, we describe our problem setting, including assumptions about the robot's environment as well as minimal specifications for its planning and learning modules.
% We note that while we make specific choices in our implementation of these modules, any other choice that respects these specifications should be compatible with our framework.

\subsection{Modelling the World}
\label{subsec:world-modelling}
\begin{figure}[t]
  \centering
    \noindent
    \includegraphics[width=\columnwidth]{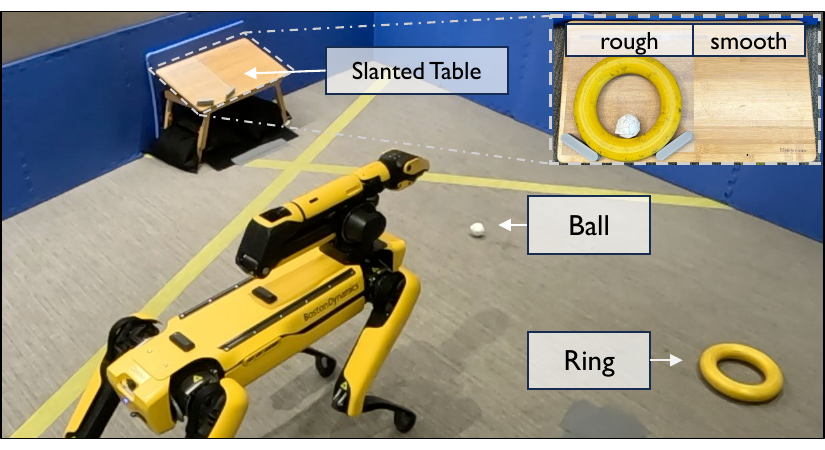}
    \vspace{-0.5em}
    \caption{\textbf{Running example: Ball-Ring environment.} The goal is to put the ball on the table. The robot should learn that (1) the ball cannot be placed directly because it will roll off the slanted table; (2) the ring can only be placed on the left side because the right side is smooth (shown in the top-right corner); (3) placing the ring on the table and then placing the ball inside the ring is the best way to accomplish the goal.}
    \vspace{-1.5em}
  \label{fig:ball-ring-env}
\end{figure}

We assume that the robot and its environment are modelled as a goal-based Markov Decision Process with object-oriented states~\cite{diuk2008object} and parameterized actions~\cite{masson2016reinforcement,ames2018learning}.
States are factored into objects and their continuous features.
For example, consider the Ball-Ring environment shown in Figure \ref{fig:ball-ring-env}.
The \texttt{ball}, \texttt{ring}, \texttt{table}, \texttt{floor}, and \texttt{robot} itself are objects, and their features include, for example, \emph{gripper joint value} (for the robot) and \emph{xyz position} (for other objects).
Of course, a real-world robot cannot perceive such features directly, so we assume that the robot is equipped with a perception system that can construct a fully-observed state $\state_t \in \states$ from raw sensory observations at each time step $t \in \mathbb{Z}^+$.
This model does not account for the perception noise that exists in the real world, but in experiments, we find that our approach is reasonably robust to that noise (Section \ref{sec:experiments}).

The action space of the MDP is defined by a set of \emph{parameterized skills} that have continuous parameters.
It is often convenient to define \emph{object parameters} as well, but for the purpose of simplifying exposition, we will treat these as part of the skill unless otherwise noted.
For example, in the Ball-Ring environment, \texttt{Place(ball},~\texttt{table},~$\circ$\texttt{)} is one skill $\parameterizedskill \in \parameterizedskills$, where $\circ$ denotes a placeholder for a continuous parameter, and \texttt{Place(ring},~\texttt{floor},~$\circ$\texttt{)} is another $\parameterizedskill' \in \parameterizedskills$.\footnote{Note that in implementation, we implement object-parameterized skills as discussed in Appendix Section \ref{appendix:experiments}.}
The continuous parameters for both skills are \emph{xy} relative offsets between the gripper and target surface (the height and gripper orientation are fixed).

To define skills formally, we use an extension of the options framework~\citep{sutton1999between}.
A parameterized skill $\parameterizedskill \in \parameterizedskills$ is given by a tuple 
$(\precondition, \Theta, \lowlevelpolicy, \termination, \successfn)$
where $\precondition: \states \to \{0, 1\}$ characterizes states where the skill can be initiated, $\Theta \subseteq \mathbb{R}^m$ is the set of possible continuous parameters, $\mu(\state, \theta)$ is a low-level controller that takes a state $\state$ and continuous parameters $\theta \in \Theta$ as input, $\termination : \states \to \{0, 1\}$ is a termination condition, and $\successfn : \states \to \{0, 1\}$ is a success condition indicating whether the skill has achieved its intended outcome in the  terminal state.
A skill with parameters assigned is treated as an atomic action.
After an action $\action_t \in \actions$ is executed, the environment advances according to an unknown transition distribution $\state_{t+1} \sim P(\cdot \mid \state_t, \action_t)$.

The robot is tasked with achieving particular \textit{goals}.
Each goal is sampled from a task distribution  $\goal \sim P(\cdot \mid \state_0)$.
For example, in the Ball-Ring environment, the goal might be that the ball is stably at rest on the table.
We do not assume direct access to the goal distribution; instead, the robot receives goals from a human gradually during learning~\ref{subsec:lifelong-learning}.
Formally, a goal is a binary classifier over states $\goal : \states \mapsto \{0, 1\}$ where 1 indicates that a state is within the goal set.
We refer to a combination of an initial state $\state_0$ and goal $\goal$ as a \textit{task}.
\emph{Solving} a task entails taking actions to reach a state $\state_t$ where $\goal(\state_t) = 1$ from $\state_0$ within a maximum time-step horizon $\horizoneval$.

\subsection{Planning to Solve Tasks}
\label{subsec:planning}

Given a task, the robot will plan to generate actions that are likely to accomplish the goal from the initial state.
Following previous work~\cite{srivastava2014combined,ames2018learning,silver2021learning,bilevel-planning-blog-post}, we decompose planning into two levels: skill sequencing and continuous parameter selection.
Skill sequencing consists of generating a \emph{skeleton}, e.g.,
$($\texttt{MoveTo(ball},~$\circ$\texttt{)},
\texttt{Pick(ball}, \texttt{floor},~$\circ$\texttt{)},
\texttt{MoveTo(table},~$\circ$\texttt{)},
\texttt{Place(ball}, \texttt{table},~$\circ$\texttt{)}$)$.
Given a skeleton, we select continuous parameters using \emph{parameter policies} $\theta \sim \parameterpolicy_\parameterizedskill(\cdot \mid \state)$.
% Note that each skill $\parameterizedskill \in \parameterizedskills$ has its own parameter policy.
Since parameter selection is conditioned on the state $\state$, and since we are not assuming a known transition distribution~\cite{silver2021learning}, we sample and execute each skill \emph{greedily}.
If the skill terminates and does not meet its success condition, we replan.
See Algorithm~\ref{alg:planning-to-solve-tasks} for a summary.

%%%%%%%%%%%%%%%%%%%%%%%%%%%%%%%%%%%%%%%%%%%%%%%%%%%%%%%%%%%%%%%%%%%%%%%%%%%%%%%%%%%%%%%%%%%%%%%%%%%%%%%
%%%%%%%%%%%%%%%%%%%%%%%%%%%%%%%%%%%%%%%%%%%%%%%%%%%%%%%%%%%%%%%%%%%%%%%%%%%%%%%%%%%%%%%%%%%%%%%%%%%%%%%

The robot will learn parameter policies through online experience (Section ~\ref{subsec:lifelong-learning}).
We assume that each skill $\parameterizedskill$ is accompanied by a \textit{parameter prior} $\parameterprior_\parameterizedskill$ to be used before any parameter policies have been learned.
For example, a pick skill may have an associated grasp sampler that provides valid grasps some percentage of the time.
At first, the robot uses the parameter priors to select parameters ($\parameterpolicy_\parameterizedskill = \parameterprior_\parameterizedskill$), but as the robot collects online experience, it will learn to improve the parameter policies with respect to the environment and task distribution.
For example, the robot should learn grasp samplers that are specialized to the objects in the environment that need to be manipulated.

How can we generate skeletons to maximize the probability that sampling will succeed?
Towards answering this question, we introduce the notion of \emph{skill competence}.

\begin{definition}[Skill Competence]
The \emph{competence} $\competence_{\parameterizedskill, \parameterpolicy}$ of a skill $\parameterizedskill$ with current parameter policy $\parameterpolicy_\parameterizedskill$ is the expected success $\mathop{\mathbb{E}}[\successfn_\parameterizedskill(\staterv_{t+1}) \mid \precondition_\parameterizedskill(\staterv_t)]$, where $\precondition_\parameterizedskill$ characterizes the skill $\parameterizedskill$ can be initiated from, $\staterv_t$ is a random variable for the state before skill execution, $\actionrv_t$ is a r.v. for the action generated from $\theta \sim \parameterpolicy_\parameterizedskill(\cdot \mid \staterv_t)$, and $\staterv_{t+1} \sim P(\cdot \mid \staterv_{t}, \actionrv_{t})$.
\end{definition}
For example, if \texttt{Pick(ball}, \texttt{floor}, $\circ$\texttt{)} successfully grasps the ball from the floor 80\% of the time, the competence would be 0.8.
Note that competence is defined in terms of the current parameter policy, and that the distribution of $\staterv_t$ is induced by the overall planning procedure and the task distribution.
In practice, skill competences are unknown and must be estimated from data (Section~\ref{sec:planning-to-practice}).

To establish a relationship between skill competence and full skeleton success, we introduce a strong assumption:
\begin{assumption}
\label{assumption:independence}
Success $\successfn_\parameterizedskill(\staterv_{t+1})$ is independent from the state $\staterv_t$ conditioned on the initiation condition $\precondition_\parameterizedskill(\staterv_t) = 1$.
\end{assumption}

In other words, the success rate of a skill is the same for all states in its initiation set.
This assumption has been previously considered in different forms~\cite{konidaris2018skills,ames2018learning}, but it does not always hold in practice.
For example, the specific grasp of an object may influence the success rate of placing. 
We can mitigate this by replanning, but to fully remove the assumption, we would need task and motion planning~\cite{srivastava2014combined,garrett2021integrated} or automated skill partitioning~\cite{konidaris2018skills,ames2018learning}, which we leave to future work.

We can now revisit the problem of generating a skeleton that has the maximum likelihood of success.
Given Assumption~\ref{assumption:independence}, we want to find a skeleton $(\parameterizedskill_0, \dots, \parameterizedskill_n)$ with three properties: (1) $\prod_{i=0}^n\competence_i$ is maximal, where $\competence_0, \dots, \competence_n$ are the corresponding competences for the skills in the skeleton; (2) the initiation and success conditions for subsequent skills chain together (see \cite{konidaris2018skills} for a formal definition); and (3) the goal is achieved.
Following previous work~\cite{ames2018learning,konidaris2018skills,silver2021learning,silver2022learning}, we take advantage of the close relationship between (parameterized) options and AI planning operators to generate skeletons that satisfy conditions (2) and (3).
Previous work has considered how to learn these operators automatically; we manually specify them for this work.
To satisfy condition (1), we associate a cost of $-\log(\competence)$ to the respective operator and use an off-the-shelf AI planner~\cite{helmert2006fast} to find a minimal cost (maximum likelihood) skeleton.
See Appendix~\ref{appendix:planning-details} for further details.

\begin{figure}[t]
\removelatexerror
\begin{algorithm}[H]
    \DontPrintSemicolon
    % \SetAlgoCaptionLayout{centerline}
    \nl \textbf{Input:} Current state $\state$ and goal $\goal$.\;
    \nl Generate a skeleton $(\parameterizedskill_0, \dots, \parameterizedskill_n)$ to $\goal$ from $\state$.\;
    \nl For $i=0, \dots, n$:\;
    \nl \quad Sample $\theta \sim \parameterpolicy_{\parameterizedskill_i}$ and \textbf{execute} $\parameterizedskill_i(\theta)$.\;
    \nl \quad Perceive and update the current state $\state$.\;
    \nl \quad If $\successfn_{\parameterizedskill_i}(\state) \neq 1$, repeat from line 2 (replan).\;
\caption{Planning and Execution}
\label{alg:planning-to-solve-tasks}
\end{algorithm}
\vspace{0.6em}

\begin{algorithm}[H]
    \DontPrintSemicolon
    % \SetAlgoCaptionLayout{centerline}
    \nl \textbf{Initialize} parameter policies $\parameterpolicies = \{ \parameterprior_\parameterizedskill : \parameterizedskill \in \parameterizedskills\}$.\;
    \nl \textbf{Repeat}:\;
    \nl \quad \textbf{If} a human has given a goal $\goal$:\;
    \nl \quad\quad Plan and execute to $\goal$ with Algorithm~\ref{alg:planning-to-solve-tasks}.\;
    \nl \quad \textbf{Else}:\;
    \nl \quad\quad Select and execute actions of the robot's choice.\;
    \nl \quad Update $\parameterpolicies$ every $m$ iterations.
\caption{Online Learning Paradigm}
\label{alg:lifelong-learning}
\end{algorithm}
\vspace{-0.8em}
\end{figure}

\subsection{Online Learning Paradigm}
\label{subsec:lifelong-learning}

We want the robot to get better at solving tasks over time.
We consider a reset-free online learning~\cite{thrun1995lifelong,lu2020reset,gupta2021reset} paradigm where the robot is sometimes given a task to solve and otherwise given \emph{free time} during which it should autonomously learn to improve.
The \textbf{key question} is: what should the robot do during free time to get better at solving tasks?

We assume that the skills themselves are fixed (e.g., for safety reasons), but the parameter policies can change.
The robot should therefore use its free time to improve its parameter policies, specializing the given parameter priors to the particular objects, goals, and constraints in its environment.
This setup is summarized in Algorithm~\ref{alg:lifelong-learning}.
Note that this setup is fully autonomous; the environment is not reset.
Our main interest is Line 6: how should the robot choose actions to gather data for improving its parameter policies?

\section{Planning to Learn}
\label{sec:planning-to-practice}

\begin{algorithm}[t]
    \DontPrintSemicolon
    % \SetAlgoCaptionLayout{centerline}
    \nl \textbf{Input}: Current parameter policies $\parameterpolicies$.\;
    \nl Select a skill $\parameterizedskill \in \parameterizedskills$ to practice (see Algorithm~\ref{alg:estimate-extrapolate-situate}). \;
    \nl Plan to $\precondition_\parameterizedskill$ using Algorithm~\ref{alg:planning-to-solve-tasks} with $\parameterpolicies$. \;
    \nl Practice the skill $\parameterizedskill$ one time: \;
    \nl \quad Sample parameters $\theta$ from an \emph{explore policy} $\exploreparameterpolicy_\parameterizedskill$.\;
    \nl \quad \textbf{Execute} $\parameterizedskill(\theta)$ and record the transition.\;
    \nl \textbf{Repeat} from line 2 until free time expires.
\caption{Planning to Practice}
\label{alg:planning-to-practice}
\end{algorithm}

We propose that the robot should spend its free time \emph{planning to practice} skills.
In particular, we commit to the meta-strategy shown in Algorithm~\ref{alg:planning-to-practice}, where the robot repeatedly selects a skill to practice, plans to satisfy that skill's initiation condition, samples parameters from an \emph{explore parameter policy}, executes the action, and records the result.
In using this meta-strategy, we make two assumptions.

\begin{assumption}
\label{assumption:reversiblity}
For $\state \in \states$ and $\parameterizedskill \in \parameterizedskills$, there exists a sequence of actions that reach $\precondition_\parameterizedskill$ from $\state$ with nonzero probability.
\end{assumption}

In other words, it is not possible to get permanently ``stuck'' during online learning.
This assumption can be weakened if certain skills do not need to be practiced infinitely often.

\begin{assumption}
\label{assumption:priorsupport}
(Informal) Parameter priors have support over good parameter choices.
\end{assumption}

For efficiency (and perhaps safety) purposes, we will not permit the robot to sample arbitrarily from skill parameter spaces; we therefore assume that the priors are sufficiently broad to enable learning.
Given these assumptions, we are left with three decisions:

\begin{enumerate}
    \item How should we decide what skills to practice?
    \item What explore parameter policies $\exploreparameterpolicy_\parameterizedskill$ should we use?
    \item How should we update the parameter policies?
\end{enumerate}

\begin{figure*}[t]
  \centering
    \noindent
    \includegraphics[width=0.98\textwidth]{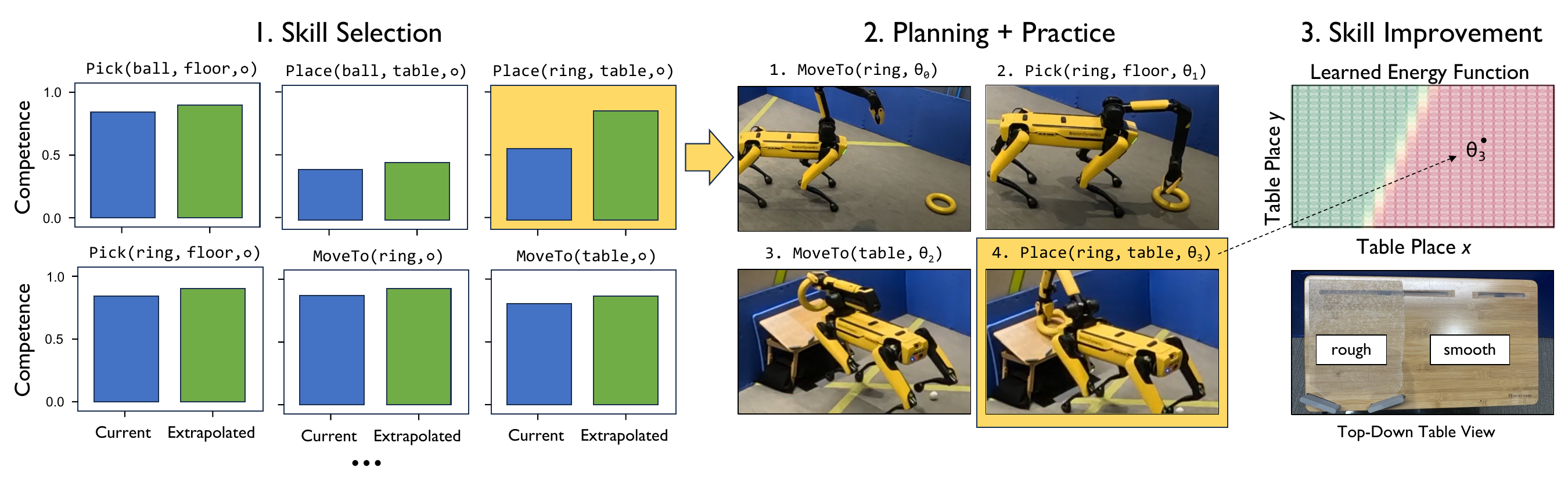}
    \vspace{-0.5em}
    \caption{\textbf{Pipeline overview.} (1) During free time, the robot repeatedly selects skills to practice. Here, \texttt{Place(ring},~\texttt{table},~$\circ$\texttt{)} is selected because it maximizes $\successfn_{\text{skill}}$ (Algorithm~\ref{alg:estimate-extrapolate-situate}). (2) The robot plans to satisfy the initiation condition of the skill and then selects a continuous parameter to practice (Algorithm~\ref{alg:planning-to-practice}). (3) The resulting success or failure of the skill is used to improve the parameter policy (Section~\ref{subsec:learning}).}
    \vspace{-1em}
  \label{fig:pipeline}
\end{figure*}

\noindent In this work, we choose to focus on the first question and draw on existing techniques to answer the second two.
See Figure~\ref{fig:pipeline} for an overview of the full pipeline.

\subsection{Selecting Skills to Practice}
\label{sec:selecting-skills-to-practice}

Given the relationship between competence and task success, a natural answer to the first question would be to practice the skill with the lowest current competence.
However, there there are two major issues with this ``Fail Focus'' strategy.
First, a low-competence skill may be impossible to improve.
For example, in the Ball-Ring environment, the \texttt{Place(ball}, \texttt{table}, $\circ$\texttt{)} skill is bound to fail since the table is slanted (as seen in Figure \ref{fig:ball-ring-env}).
Second, even if a low-competence skill could be improved, the skill may be less critical for the task distribution than others.
In the worst case, Fail Focus may cause the robot to spend all its free time attempting to improve an impossible skill that is of no consequence to any given task.

A better skill selection strategy would be more directly tied to our real objective: to efficiently and effectively solve the tasks given to the robot.
We consider a close proxy to this real objective.
Given parameter policies $\parameterpolicies = \{\parameterpolicy_\parameterizedskill : \parameterizedskill \in \parameterizedskills\}$ and a task $(\state_0, \goal)$, let $\successfn_{\text{task}}(\parameterpolicies, \state_0, \goal)$ be the probability that planning succeeds without replanning.
From Section~\ref{subsec:planning}, we have that $\successfn_{\text{task}}(\parameterpolicies, \state_0, \goal) = \prod_{i=0}^n \competence_i$, where $\competence_i$ is the competence of the $i^{\text{th}}$ skill in the skeleton generated for the task.
Given a task distribution $(\staterv_0, \goalrv)$, our overall objective is to learn parameter policies that maximize:
\begin{equation}
    \successfn_{\text{tasks}}(\parameterpolicies) \triangleq \mathop{\mathbb{E}}_{\staterv_0, \goalrv}[\successfn_{\text{task}}(\parameterpolicies, \staterv_0, \goalrv)],
\label{eq:objective}
\end{equation}
that is, the expected probability that planning succeeds without replanning over the task distribution.

We propose to practice the skill whose predicted improvement would maximally increase $\successfn_{\text{tasks}}(\parameterpolicies)$.
In other words, we will practice \textbf{the skill with the greatest expected improvement to the overall distribution of human-given tasks}.
We do this in three key steps (Algorithm~\ref{alg:estimate-extrapolate-situate}).
For each skill, we:
(1) \textit{Estimate the competence}: compute the current competence of the skill from data;
(2) \textit{Extrapolate the competence}: predict how the competence of the skill would change if it were practiced once more and then its parameter policy was updated;
(3) \textit{Situate the competence}: predict how the overall success rate on the task distribution would change given the extrapolated skill competence.
This approach resolves the issues with Fail Focus: by extrapolating, we avoid practicing impossible or plateaued skills; and by situating, we avoid practicing irrelevant or unimportant skills.
We now describe these steps in detail.

\begin{algorithm}[t]
    \DontPrintSemicolon
    % \SetAlgoCaptionLayout{centerline}
    % \nl \textbf{Input}: Current parameter policies $\parameterpolicies$.\;
    \nl For each $\parameterizedskill \in \parameterizedskills$ with current parameter policy $\parameterpolicy_\parameterizedskill$: \;
    \nl \quad \textbf{Estimate} the current competence $\competence_{\parameterizedskill,\parameterpolicy}$.\;
    \nl \quad \textbf{Extrapolate}: predict $\competence_{\parameterizedskill, \parameterpolicy'}$, the competence after \\ \vspace{-0.075em} \quad practicing $\parameterizedskill$ and updating $\parameterpolicy_\parameterizedskill$ to $ \parameterpolicy_\parameterizedskill'$.\;
    \nl \quad \textbf{Situate} the competence in the task distribution, \\ \quad   computing $\successfn_{\text{skill}}(\parameterizedskill) \triangleq \successfn_{\text{tasks}}(\parameterpolicies - \{\parameterpolicy\} \cup \{\parameterpolicy'\})$.\;
    \nl \textbf{Return} $\argmax_{\parameterizedskill} \successfn_{\text{skill}}(\parameterizedskill)$ for practice.
\caption{Selecting a Skill to Practice}
\label{alg:estimate-extrapolate-situate}
\end{algorithm}

%%%%%%%%%%%%%%%%%%%%%%%%%%%%%%%%%%%%%%%%%%%%%%%%%%%%%%%%%%%%%%%%%%%%%%%%%%%%%%%%%%%%%%%%%%%%%%%%%%%%%%%

\subsubsection{Estimating Skill Competence}
Our first task is to estimate the current competence of a skill based on the transitions that have been collected thus far.
To estimate competence, we propose a graphical model that explicates the relationship between competence, transitions, and learning.

Recall that parameter policies are updated periodically (Algorithm~\ref{alg:lifelong-learning}); we refer to each period with the same parameter policy (i.e., before an update is made) as a \emph{learning cycle}.
Thus, the robot's free time is composed of a series of learning cycles.
Let $\outcome_{i,k}$ be a binary random variable for the result of the success condition on the $k^{th}$ usage of the skill in learning cycle $i$.\footnote{We have not yet defined the explore policy, but it is important here that only ``exploit'' samples are used to estimate competence; see Section ~\ref{subsec:explore-parameter-policies}.}
Note that different skills are selected for practice each cycle, so the number of skill usages (maximum $k$ values) varies and can be zero.
In practice, this number is low, so we cannot reliably estimate the competence of a skill based on data from the current cycle alone.

Let $\competencerv_i$ be a random variable for the skill's competence \emph{during} learning cycle $i$ (before it has re-learned using the data from that cycle).
To estimate skill competence, we wish to know $P(\competencerv_{k} \mid \outcome_{1,1}, \outcome_{1,2}, \dots, \outcome_{1,n_{1}}, \outcome_{2,1}, \dots, \outcome_{k,n_{k}})$, that is, the conditional distribution of competence \emph{now} given all observations up until now.
We consider a Bayesian time series model with a joint distribution that factorizes as follows:
$$P(\outcome_{1:n_{1}}, ..., \outcome_{k:n_{k}}, ..., \competencerv_{1:N_{\text{cycle}}}) = \prod_i P_{i}(\competencerv_{i})\prod_k P(\outcome_{i,k} \mid C_i)$$
where $P(\outcome_{i,k} \mid \competencerv_i)$ is the \emph{observation model} and $P_{i}(\competencerv_{i})$ is a \emph{cycle prior}.
For the observation model, we use a Bernoulli:
$$P(\outcome_{i,k} = 1 \mid \competencerv_i = \competence) = \competence.$$

For the cycle priors, we use Beta distributions, since the Beta is the conjugate prior of the Bernoulli.
Let us first consider $P_0(\competencerv_0)$, the prior competence before any observations have been made.
We define one prior for all skills that has high mean but large variance (in experiments, $\text{Beta}(10, 1)$), reflecting our weakly held expectation that parameter priors $\parameterprior_\parameterizedskill$ will be generally good.
This also introduces a form of optimism, which can be helpful for exploration~\cite{auer2002using,may2012optimistic}.

For subsequent cycle priors $P_i(\competencerv_i)$, we assume that a skill's competence is some function of the size of the dataset used to learn that skill's parameter policy.
Note that we are \emph{not} positing a \emph{general} relationship between data count and competence.
One skill may always have perfect competence; another may always have zero competence; and a third may improve as data increases.
Let $f_\phi : \mathbb{Z}_{\ge 0} \to \text{Beta}(\alpha, \beta)$ be a \emph{competence model} where the input is the number of data used for learning, the output is a Beta distribution over competence and $\phi$ indicates that the function belongs to a hypothesis class $\mathcal{F}$.
The cycle prior $P_i(\competencerv_i)$ is given by $f_\phi(m_i)$ where $m_i$ is the number of data collected for the skill through cycle $i$.
For example, in Ball-Ring, a good competence model for the \texttt{Place(ball}, \texttt{table},~$\circ$\texttt{)} skill would output a near-zero Beta for any input, because no amount of data can improve the skill.
On the other hand, a competence model for \texttt{Place(ring}, \texttt{table},~$\circ$\texttt{)} should output Beta distributions with increasing modes, since that skill \emph{can} improve with practice.

To estimate the current skill competence, we need to infer  $\competencerv_i$ for all $i$ and fit $\phi$ for $f_\phi$.
We considered two approaches: a principled expectation-maximization (EM) approach, and a much simpler sliding-window-based approach.
In preliminary experiments, we found the simpler approach to perform at least as well as EM, and its behavior was much easier to interpret, so we used it for our main experiments.
See Appendix \ref{appendix:comp-estim-details}.

\subsubsection{Extrapolating Skill Competence}

Given the competence model $f_\phi$ fit during estimation, extrapolation is straightforward: we can simply evaluate $f_\phi(m + 1)$ to predict how the skill competence would change if we collected one more data point of practice, where $m$ is the number of data seen so far.
Let $\hat{\competence}$ denote the mode of $f_\phi(m + 1)$, i.e., the most likely next competence. 
Here we assume that a skill's competence never gets worse with learning: $\forall \phi$, if $m' > m$, then $\mathbb{E}[f_\phi(m')] \ge \mathbb{E}[f_\phi(m)]$.
We can enforce this assumption by choosing $\mathcal{F}$ appropriately.
This assumption may not always hold in practice, but for the purpose of extrapolation, it is important that the agent be optimistic and not deliberately avoid collecting additional data for a skill.

\subsubsection{Situating Skill Competence}

Our final step is to predict the expected improvement to the overall task distribution given the extrapolated competence.
Let $\parameterpolicies'$ be the set of current parameter policies $\parameterpolicies$, but with the parameter policy $\parameterpolicy_\parameterizedskill$ for the current skill under consideration $\parameterizedskill$ replaced with $\parameterpolicy'_{\parameterizedskill}$, a hypothetical policy that would result from practicing $\parameterizedskill$ once more and re-learning.
We wish to compute $\successfn_{\text{tasks}}(\parameterpolicies')$ (Equation~\ref{eq:objective}), the expected probability that planning would succeed (without replanning) over the task distribution.

To compute $\successfn_{\text{tasks}}(\parameterpolicies', \state_0, \goal)$ for a given task $\state_0, \goal$, we need not know $\parameterpolicy'_{\parameterizedskill}$ itself, but only the competence of $\parameterpolicy'_{\parameterizedskill}$, which we have computed by extrapolating.
To complete $\successfn_{\text{tasks}}(\parameterpolicies')$, we need to take an expectation over the task distribution.
As mentioned in Section \ref{sec:problem-setting}, we do not assume that the robot has direct access to the task distribution; instead, we collect the states and goals used to query the planner, including when replanning is triggered (Algorithm~\ref{alg:planning-to-solve-tasks}) and use that empirical task distribution to approximate $\successfn_{\text{tasks}}(\parameterpolicies')$.

\subsection{Explore Parameter Policies}
\label{subsec:explore-parameter-policies}

After we have selected a skill to practice and planned to satisfy its initiation condition, we must decide what parameters to use (Algorithm~\ref{alg:planning-to-practice}).
We can view this parameter selection problem as a contextual bandit with infinite arms \citep{bandits,li2010contextual}.
Here, the context is the current state $\state_t$, the actions are parameters $\theta$, and the reward is 1 if the success condition passes and 0 otherwise.
To balance exploration and exploitation, we use an epsilon-greedy policy:
$$\exploreparameterpolicy_\parameterizedskill = \epsilon\parameterprior_\parameterizedskill + (1 - \epsilon)\parameterpolicy_\parameterizedskill.$$
Other choices are possible; we use this simple approach to maintain focus on the skill selection problem.

\subsection{Learning to Improve Parameter Policies}
\label{subsec:learning}

To complete our approach for planning to practice parameterized skills, we must now determine how the collected experience can be used to improve the parameter policies.
Recall that for each skill $\parameterizedskill$, we are given a parameter prior $\parameterprior_\parameterizedskill$ and we wish to learn an improved parameter policy $\parameterpolicy_\parameterizedskill$.
Recall also that we have recorded transition data, which we can partition by skill and label according to whether the $\successfn$ check for that particular skill passed: $\dataset_\parameterizedskill = \{((\state_t, \action_t),\successfn_\parameterizedskill(\state_{t+1})) : \action_t \text{ uses skill } \parameterizedskill\}$.
Given these data, many approaches are possible.
Our approach is to learn an implicit (energy) function $\energyfn_\parameterizedskill: \states \times \actions \to \mathbb{R}_{\ge 0}$ to define the parameter policy:
$$\parameterpolicy_\parameterizedskill(\theta \mid \state) \propto \parameterprior_\parameterizedskill(\theta \mid \state)\energyfn_\parameterizedskill(\state, \parameterizedskill(\theta)).$$
Specifically, we train small neural network classifiers to minimize binary cross entropy loss and then use the classifier log probabilities for $\energyfn_\parameterizedskill$.
In contrast to the explore policy, the parameter policy $\parameterpolicy_\parameterizedskill$ is meant to exploit, so we select parameters via $\argmax_\theta{\parameterpolicy_\parameterizedskill(\theta \mid \state})$.
In practice, we sample 100 candidates from the prior $\parameterprior_\parameterizedskill(\theta \mid \state)$ and select the maximum.

For real-robot experiments, it is essential that parameter policies can be learned from very little data.
Beyond using good parameter priors, we take two additional steps for data efficiency.
First, we share neural network weights between parameter policies that have the same ``parent'' skill but different object parameters (e.g., \texttt{Place(ball},~\texttt{table},~$\circ$\texttt{)} and \texttt{Place(ring},~\texttt{floor},~$\circ$\texttt{)}).
Second, we perform feature engineering for learning by mapping the full state and action to a low-dimensional vector that is input to $\energyfn_\parameterizedskill$.
After applying these features to the data in $\dataset_\parameterizedskill$, we are left with a standard binary classification dataset.
See Appendix \ref{appendix:approach-details} for details of the feature mapping as well as the training and use of $\energyfn_\parameterizedskill$.

\section{Experiments}
\label{sec:experiments}
Our experiments are designed to empirically answer the following questions about our approach (EES):
\vspace{1em}
\begin{tightlist}
    \item[\textbf{Q1.}] To what extent does EES choose skills for practice that lead to improvements in task distribution success rate, especially compared to alternative approaches?
    \item[\textbf{Q2.}] How sample efficient is EES compared to alternatives?
    \item[\textbf{Q3.}] To what extent is EES aware of the task distribution?
\end{tightlist}
\vspace{1em}

%%%%%%%%%%%%%%%%%%%%%%%%%%%%%%%%%%%%%%%%%%%%%%%%%%%%%%%%%%%%%%%%%%%%%%%%%%%%%%%%%%%%%%%%%%%%%%%%%%%%%%%
\begin{figure*}[t]
  \centering
    \noindent
    \includegraphics[width=0.98\textwidth]{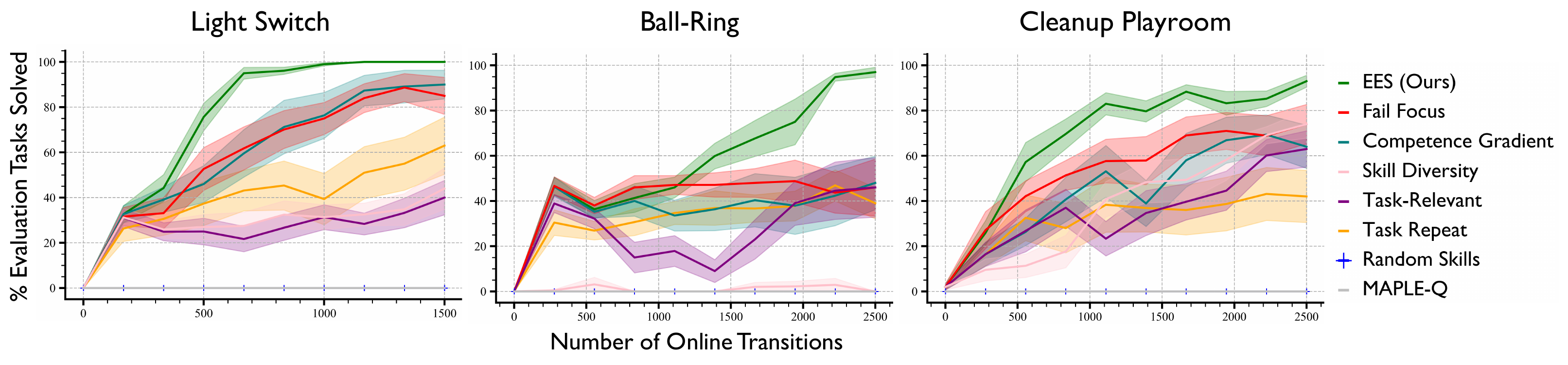}
    \caption{\textbf{Simulation results.} Percentage of evaluation tasks solved vs. number of online transitions collected for all approaches in all simulated environments. Solid lines represent means and shading represents standard error across 10 seeds. Note that all approaches used the same parameter priors and feature engineering, discussed in detail in Appendix Sections \ref{appendix:experiments} and \ref{appendix:param-policy-details}. We run additional ablation experiments on these choices in Appendix Section \ref{appendix:additional-experiments}.}
  \label{fig:sim-results}
\end{figure*}
%%%%%%%%%%%%%%%%%%%%%%%%%%%%%%%%%%%%%%%%%%%%%%%%%%%%%%%%%%%%%%%%%%%%%%%%%%%%%%%%%%%%%%%%%%%%%%%%%%%%%%%

\textbf{Environments.}
We now provide high-level environment descriptions with details in Appendix \ref{appendix:experiments}.
We use three simulated environments of varying complexity and two real-robot analogs to simulated environments.
For details on our real-robot setup, see Appendix \ref{appendix:robot-details}.
See also the supplementary material for time-lapse videos of the real robot practicing skills and learning over time.
\begin{tightlist}
    \item \textit{Light Switch (Simulated)}: A toy 1D grid environment. The robot starts in the leftmost room (grid cell) and must switch on a light in the rightmost room. The light is controlled by a dial in the same room that must be precisely actuated. The robot has skills to move left or right, turn the dial to a sampled setting, and also try to ``jump'' from a particular room all the way to the final room with the dial, though this jump skill is impossible and always fails. We use a grid size of 25 rooms in our main experiments.
    \item \textit{Ball-Ring (Simulated)}: A simulated version of the environment depicted in Figure~\ref{fig:ball-ring-env}. To add complexity, the simulated version features multiple tables, some of which are slanted and partially smooth (as shown in the figure) and others that are standard flat surfaces. The relative locations of the smooth patches vary between tables. The robot does not initially know that placing the ball on a slanted surface will fail, nor does it know that placing the ring on the smooth part of a slanted table will fail with high probability. The robot has skills for moving, picking, and placing the ball and ring.
    \item \textit{Ball-Ring (Real)}: The real version of the previous environment. See Figure~\ref{fig:ball-ring-env} and the supplementary video.
    \item \textit{Cleanup Playroom (Simulated)}: A simulated version of the environment depicted in Figure~\ref{fig:teaser}. The robot is tasked with cleaning up a child's playroom by putting two toys into a bin. The toys start out atop a table that may be blocked by a chair. There is also a brush on the floor. The robot again has skills for moving, picking, and placing, but also for grasping and dragging the chair, grasping and dumping out the bin, and sweeping toys from the table into the bin. The success of sweeping depends on the relative positions of the toys and bin and the sweeping velocity.
    \item \textit{Cleanup Playroom (Real)}: The real version of the simulated Cleanup Playroom environment. See Figure~\ref{fig:teaser} and the supplementary video.
\end{tightlist}

\textbf{Approaches.}
We now briefly describe all of the approaches that we evaluate.
See Appendix~\ref{appendix:approach-details} for details.
The first five approaches are alternative instantiations of \emph{planning to practice} (Algorithm~\ref{alg:planning-to-practice}); the last three do not use that meta-strategy.
\begin{tightlist}
    \item \textit{Estimate, Extrapolate, Situate (EES)}: Our main approach.
    \item \textit{Fail Focus}: Practices the skill with the lowest current estimated competence.
    \item \textit{Competence Gradient}: Inspired by~\citep{stout2010competence,baranes2013active,colas2019curious}, practices the skill with the highest expected competence gain (i.e., difference between current and extrapolated competence).
    \item \textit{Skill Diversity}: Practices the least-practiced skill.
    \item \textit{Task-Relevant}: Practices a randomly selected skill among those that have been previously included in some maximum-likelihood task plan.
    \item \textit{Task Repeat}: Samples a task from the empirical task distribution and plans to the task goal. If the task goal is already reached, plans instead to the task initial state.
    \item \textit{Random Skills}: Repeatedly executes random skills selected from those whose initiation conditions are satisfied in the current state. Does not plan.
    \item \textit{MAPLE-Q}: Uses hierarchical reinforcement learning to learn to select both skills and parameters. Does not plan. This approach is a modified version of MAPLE \citep{nasiriany2022maple}, which was developed for a more difficult problem setting. Our modifications are meant to give the approach access to the same prior knowledge as the other approaches (e.g., explicit option definitions and parameter priors), but a direct comparison remains challenging. See Appendix \ref{appendix:approach-details} for further discussion and details.  
\end{tightlist}

\begin{table}[t]
\resizebox{\columnwidth}{!}{%
\begin{tabular}{@{}ccccccc@{}}
\toprule
\textbf{\# Steps} & \textbf{S1} & \textbf{S2} & \textbf{S3} & \textbf{S4} & \textbf{S5} & \textbf{Mean} \\ \midrule
0 & 0.0 & 0.0 & 0.0 & 0.0 & 0.0 & \textbf{0.0} \\
120 & 0.0 & 0.33 & 0.0 & 0.67 & 0.0 & \textbf{0.2} \\
240 & 1.0 & 0.67 & 1.0 & 0.67 & 0.67 & \textbf{0.8} \\ \bottomrule
\end{tabular}%
}
\caption{\textbf{Ball-Ring real-robot results.} Steps are online transitions; S1 = Seed 1. Entries are fractions of 3 evaluation tasks solved.}
\label{tab:brt-playroom-results}
\end{table}

\textbf{Experimental Setup.} For all simulated environments, we run 10 random seeds of each approach.
Task generation, parameter prior sampling, tie-breaking during planning, and stochastic environment transitions all vary between seeds.
Within each environment, the number of free periods, the number of steps within each free period, and the evaluation horizon $\horizoneval$ is the same for all approaches (see Appendix~\ref{appendix:experiments}).
After every free period, each approach is evaluated on 10 randomly sampled ``evaluation tasks'' with held-out initial states.
All simulation experiments were conducted on a quad-core Intel Xeon Platinum 8260 processor with a 192GB RAM limit.
For real-robot environments, we run our main approach only (due to the intense time and resource requirements) and use 5 random seeds.
Each seed represents a fully independent learning trial.
We load model checkpoints after learning and use 3 tasks to evaluate performance.
For all experiments, our key quantitative measure is the robot's task success rate under the evaluation horizon.

\begin{table}[t]
\resizebox{\columnwidth}{!}{%
\begin{tabular}{@{}ccccccc@{}}
\toprule
\textbf{\# Steps} & \textbf{S1} & \textbf{S2} & \textbf{S3} & \textbf{S4} & \textbf{S5} & \textbf{Mean} \\ \midrule
0 & 0.0 & 0.67 & 0.33 & 0.67 & 1.0 & \textbf{0.53} \\
120 & 0.67 & 1.0 & 1.0 & 1.0 & 1.0 & \textbf{0.93} \\
240 & 1.0 & 1.0 & 0.67 & 1.0 & 1.0 & \textbf{0.93} \\ \bottomrule
\end{tabular}%
}
\caption{\textbf{Cleanup Playroom real-robot results.} See Table~\ref{tab:brt-playroom-results} caption for details.}
\label{tab:sweep-playroom-results}
\end{table}

\textbf{Results and Analysis.} Figure \ref{fig:sim-results} shows a plot of the evaluation success rate of all approaches in all simulated environments.
EES is consistently the most sample efficient, achieving higher success rates after fewer online transitions than the baselines.
This is particularly evident when the number of online transitions is high enough for EES to practice useful skills a sufficient number of times for a noticeable improvement.
Fail Focus falters not only because it focuses on impossible skills, like jumping in Light Switch or placing the ball on the table in Ball Ring, but also because it is not sufficiently situated in the task distribution.
For example, there are no impossible skills in Cleanup Playroom, but EES still outperforms Fail Focus.
Competence Gradient is competitive with Fail Focus but similarly lacks situatedness.
The poor performance of Skill Diversity, Task-Relevant, Task Repeat, and Random Skills underscore the importance of directed active practice in these long-horizon environments with many possible skills.
Like the Random Skills baseline, MAPLE-Q fails to solve any evaluation tasks, which is not surprising given the highly limited number of online transitions that we are considering; previous related work~\cite{nasiriany2022maple} learns from multiple orders of magnitude more data.
We verified that our implementation does well in far simpler environments with far more data; see Appendix \ref{appendix:approach-details} for further discussion.

Tables~\ref{tab:brt-playroom-results} and \ref{tab:sweep-playroom-results} show EES evaluation results for the real-robot environments.
Our approach is able to improve its performance after 120 and 240 real-world skill-executions respectively.
Each seed took 1-3 hours of real robot time; see the supplementary video.
These results are especially noteworthy given the complex skills (e.g. sweeping from the table), relatively long-horizons necessary for both setting up practice and for solving each task, and non-trivial noise in perception and control, which lead to both false positives and false negatives in the skill datasets ($\dataset_\parameterizedskill$) collected for parameter policy learning.

Towards answering Q3 (to what extent is EES aware of the task distribution?), we conducted an additional experiment in the simulated Cleanup Playroom environment.
Rather than giving the robot the goal of putting \emph{both} toys into the bin, we instead gave the goal of putting \emph{one} toy in the bin.
As shown in Table~\ref{tab:skill-practice-frequencies}, this task distribution change led to a corresponding change in practicing behavior: rather than practicing sweeping, the robot instead practiced \emph{dropping} the toy into the bin directly.
This finding matches intuition: sweeping is unnecessarily complicated when only one object needs to be stowed (a pick-and-place strategy is better), but worthwhile when two objects can be stowed with one sweep.

\begin{table}[t]
\resizebox{\columnwidth}{!}{%
\begin{tabular}{@{}cccc@{}}
\toprule
\normalsize
{\textbf{Goals}} & \normalsize{\texttt{Pick(brush)}} & \normalsize{\texttt{Drop(toy, bin)}} & \normalsize{\texttt{Sweep(...)}}\\ \midrule
\normalsize{Both toys} & \normalsize{217.1} & \normalsize{0.0} & \normalsize{110.2} \\[0.25em]
\normalsize{One toy only} & \normalsize{88.4} &  \normalsize{305.9} & \normalsize{0.0} \\ \bottomrule
\end{tabular}%
}
\caption{\textbf{Skills practiced with varying task distributions.} The entries are the total number of times EES chooses to practice a select number of skills in the simulated Cleanup Playroom environment (averaged over 10 seeds). See text for discussion.}
\label{tab:skill-practice-frequencies}
\end{table}

\section{Related Work}
\label{sec:related-work}

% Please make sure to include \verb!natbib.sty! and to use the
% \verb!plainnat.bst! bibliography style. \verb!natbib! provides additional
% citation commands, most usefully \verb!\citet!. For example, rather than the
% awkward construction 

% We now detail the relationship between this work and previous related work.

\subsection{Exploration in Reinforcement Learning}

The problem of sequentially selecting actions that lead to efficient learning is central to \emph{exploration} in reinforcement learning~\cite{kearns2002near,pathak2019selfsupervised,colas2019curious,bougie2020skill,amin2021survey}.
One important difference between our setting and RL is that we do not have a temporal credit assignment problem: given skill success conditions and Assumption~\ref{assumption:independence}, each parameter policy learning problem is self-contained.
Our skill selection problem is therefore related to exploration in multi-armed bandits~\cite{bubeck2009pure}, but different still, since selecting a skill to practice does not lead to a task reward, but rather, to a data point that can be used to improve the parameter policy for that skill.
Note that the inner problem of choosing parameters for a selected skill \emph{is} a bandit problem (with infinite arms)~\cite{carpentier2015simple} as explained in Section \ref{subsec:explore-parameter-policies}, but our primary interest is the outer skill selection problem.
Our method for skill selection can also be viewed as a restricted form of curriculum learning for RL \cite{narvekar2020curriculum}.

Within the RL literature, the most related work to ours is that of \citet{stout2010competence}, \citet{baranes2013active} and \citet{colas2019curious}, who each consider a form of \emph{competence progress} to guide exploration.
Compared to our approach, these previous works \emph{estimate} and \emph{extrapolate} competence, but they do not \emph{situate} the competence in a task distribution.
Their motivation is different from ours---they assume that a task distribution is not known and consider the problem of deriving intrinsic motivation in the absence of goals~\cite{santucci2020intrinsically}.
The Competence Gradient baseline in our experiments is inspired by these works and confirms the importance of situating competence in our setting.

Recent work by \citet{vats2023efficient} considers a variation of estimation, extrapolation, and situation in the context of \emph{recovery learning}, where additional skills are learned to supplement a given set of imperfect parameterized skills.
Instead of asking which \emph{parameter policy} to practice, they consider which \emph{recovery skill} to practice.
They introduce a ``value of failures'' objective that is similar to our $\successfn_{\text{tasks}}$ objective.
One important difference is that they assume access to a simulator; they therefore do not need to \emph{plan to practice}.

\subsection{Parameter Policy Learning in RL}

The problem of learning skill parameter policies has also been considered in the RL literature, for example, in Parameterized Action MDPs (PAMDPs)~\cite{hausknecht2015deep,masson2016reinforcement,dalal2021accelerating,nasiriany2022maple}.
In addition to facing the challenge of temporal credit assignment, these works typically do not assume a given method for discrete skill sequencing and instead need to learn a high-level ``manager'' policy in addition to the ``worker'' parameter policies.
Altogether, this represents a much harder problem setting than ours, and the sample complexity of current techniques remains prohibitively high for the kind of rapid skill specialization we consider here (for example, see our MAPLE-Q baseline~\cite{nasiriany2022maple}).
Assuming that the environment can be automatically reset is the norm in this literature, with some notable exceptions~\cite{thrun1995lifelong,lu2020reset,gupta2021reset}.
Work by \citet{ames2018learning} is a step toward bridging the RL setting and our setting; they automatically derive AI planning operators from parameterized option specifications.
Exploration is not a central consideration in these works (but see~\cite{da2014active,andersen2017active}).

Recent work in RL also considers specializing (fine tuning) skills through online learning~\cite{gupta2023bootstrapped,xiong2024adaptive}, starting from generic (pre-trained) distributions.
Other recent work by \citet{fang2022planning} is another example of \emph{planning to practice}.
Their planning uses a learned latent subgoal space, rather than options and AI planners.
They learn goal-conditioned policies that are analogous to our parameter policies.
Active learning at the skill level is not a primary focus; their approach is similar to our Task Repeat baseline (but with environment resets instead of reset-free learning).
Follow-up work~\cite{fang2023active} considers active learning more centrally and proposes a method for generating subgoals for online learning using a diversity-based metric.

\subsection{Learning Samplers for Task and Motion Planning}

In the context of the task and motion planning (TAMP) literature~\cite{garrett2021integrated}, our parameter policies can be seen as \emph{samplers} for refining skeletons generated by task planning.
TAMP approaches typically do not make Assumption~\ref{assumption:independence} and instead sample parameters contingent on the entire skeleton (e.g., selecting grasp parameters that enable future constrained placements).

Several works have considered learning samplers for TAMP~\cite{ltampkim2,ltampchitnis}.
\citet{silver2022learning} learn samplers from an offline demonstration dataset.
The details of our neural-network learning over object-centric states are most similar to theirs.
Other recent work has considered learning samplers with diffusion models~\cite{mishra2023generative,yang2023diffusion}.
Most relevant of these is the work by~\citet{mendez2023embodied}, who consider diffusion-based sampler learning for TAMP in an \emph{embodied lifelong} setting.
However, in that work, the robot is not given free time; it remains in task time throughout online learning. Additionally, the agent does not have separate exploration and exploitation samplers, but rather only an exploitation sampler.
\citet{wang2021learning} consider active sampler learning for TAMP with a focus on the inner bandit problem of selecting parameters to practice for a given skill.
In principle, their parameter selection method could be swapped in for our epsilon-greedy approach.

In the TAMP literature, the work by~\citet{noseworthy2021active} is another instance of active learning.
They learn to predict whether a skeleton is feasible~\cite{wells2019learning,driess2020deep,xu2022accelerating}, i.e., whether there exists continuous parameters that would achieve the goal (typically in a deterministic setting).
Future work could combine active feasibility prediction with our active parameter policy learning as a path toward removing Assumption~\ref{assumption:independence} and scaling to more geometrically complex environments.

\section{Limitations and Future Work} 
\label{sec:conclusion}

In this work, we proposed Estimate, Extrapolate \& Situate (EES) as a method for planning to practice parameterized skills.
We found that simulated and real robots using EES are able to rapidly and continually improve their parameter policies with respect to human-given task distributions.
Our real-robot results are particularly noteworthy as instances of reset-free online learning in challenging, long-horizon mobile manipulation environments.

There are several limitations of the present work and of EES as a general method.
For the sake of rapidly learning on a real robot, we started with a considerable amount of prior knowledge: known object (feature) detectors, fully-specified parameterized skills (and operators for planning), low-dimensional feature selectors for parameter policy training, and good parameter priors.
Previous work has considered learning each of these components (e.g., ~\cite{silver2022learning,singh2021parrot}), but doing so may require significantly more data than what we considered here.
We also made Assumptions 1-3, which are strong, and while we need not satisfy them completely to attain good performance, they remain worthy of further scrutiny.
Our approach also implicitly assumes the overall task distribution is stationary, which may not hold in many complex and dynamic real-world environments.
Our relatively naive treatment of noise and our assumption of full observability are also clearly limiting.
Furthermore, our commitment to \emph{planning to practice} (Algorithm~\ref{alg:planning-to-practice}) is perhaps overly myopic: better strategies might anticipate that practicing one skill enables quickly practicing another, reasoning over \emph{sequences} of practice attempts.
Finally, this work presupposes that robots \emph{should} be practicing and learning during deployment.
The extent to which this is true depends greatly on the nature of the deployment and the constraints under which the robot is allowed to practice.

One future direction that could address multiple limitations simultaneously would be to give the agent access to a simulator.
A significant challenge with doing this is that the precise models of all the objects the robot might encounter during deployment are not available ahead of time, and thus the robot must acquire aspects of these online.
However, even if the acquired simulation were a coarse approximation of the real world, the robot could nonetheless use it to bootstrap real-world practice time.
The same simulation could be used for reasoning about potentially irreversible actions before executing them in the real world, and for integrated task and motion planning (TAMP)~\cite{garrett2021integrated} towards removing Assumption~\ref{assumption:independence}.
Leveraging TAMP would also be a step toward more principled planning in stochastic~\cite{shah2020anytime} and partially-observable environments~\cite{garrett2020online}.

\section*{Acknowledgments}
We gratefully acknowledge support from NSF grant 2214177; from AFOSR grant FA9550-22-1-0249; from ONR MURI grant N00014-22-1-2740; from ARO grant W911NF-23-1-0034 and from the MIT Quest for Intelligence.
Nishanth, Tom, and Willie are supported by NSF GRFP fellowships.
We thank  Will Shen for feedback and suggestions on an early paper draft, as well as invaluable help with creating the accompanying website.
We also thank Russell Mendonca for helpful early discussions, especially with respect to the MAPLE-Q baseline.
We are grateful for helpful discussion, brainstorming, and support from Stefanie Tellex, Ashay Athalye, Tushar Kusnur, Jiuguang Wang, Gustavo Goretkin, Andrew Messing, Joe St. Germain and others at the AI Institute.
We thank Hannah Blumberg for invaluable early help with prototyping one of our robot domains, as well as helpful comments on an early draft of this paper.
We also thank Chris Agia, Shivam Vats, and Peter Stone for helpful comments on an earlier version of the paper.
We gratefully acknowledge the MIT SuperCloud and Lincoln Laboratory Supercomputing Center for providing HPC resources that have contributed to the simulation results reported within this paper.
Finally, we wish to thank our three Spot robots, Moana, Donner, and Kepler, for being so reliable throughout the extensive prototyping and experimentation required for this paper.
Any opinions, findings, and conclusions or recommendations expressed in this material are those of the authors and do not necessarily reflect the views of our sponsors.

%% Use plainnat to work nicely with natbib. 

\bibliographystyle{plainnat}
\bibliography{references}

\newpage
\clearpage
\appendix
% Note: for drafting purpose, TODO list
% Note: this will be automatically removed if using "disabled" in "\usepackage[disable,textsize=tiny]{todonotes}"
% \listoftodos

% %%%%%%%%%%%%%%%%%%%%%%%%%%%%%%%%%%%%%%%%%%%%%%%%%%%%%%%%%%%%%%%%%%%%%%%%%%%%%%%%%%%%%%%%%%%%%%%%%%%%%%%
% \begin{figure*}[t]
%   \centering
%     \noindent
%     \includegraphics[width=\textwidth]{figures/planning-skill-competence-improvement.pdf}
%     \caption{\textbf{Skill Selection via Task Competence Improvement for a single goal.} (Left Panel) three example skills - ground with discrete parameters - as well as graphs showing their current and extrapolated competence at a particular timestep during learning. (Right Panel) illustration of the most-likely plan for achieving the goal \texttt{On(ring, table)}. The bottom-most skill in the left panel (\texttt{Place(ball, table)}) is not used as part of the most-likely plan for the goal in question, and also has no projected competence improvement. Thus, we give it a very low score. The middle skill in the left panel (\texttt{Pick(ball, floor)}) is used as part of the most-likely plan, but has a relatively low projected competence improvement. The top skill in the left panel (\texttt{Place(ring, table)}) is used in the most-likely plan and has a relatively high projected competence improvement, and thus gets a high score as a skill to practice.}
%     % \vspace{-1.75em}
%   \label{fig:placeholder}
% \end{figure*}
% %%%%%%%%%%%%%%%%%%%%%%%%%%%%%%%%%%%%%%%%%%%%%%%%%%%%%%%%%%%%%%%%%%%%%%%%%%%%%%%%%%%%%%%%%%%%%%%%%%%%%%%

\subsection{AI Planning Details}
\label{appendix:planning-details}
In this section, we provide additional details about AI planning to supplement the overview given in Section~\ref{subsec:planning}.
We refer the reader to other references~\cite{helmert2006fast} for a formal treatment of the planning techniques we use in this work and give a brief overview of the salient points here.
First, we assume access to a set of \emph{predicates} and a function $\abstractfn$ that maps states to sets of ground predicates.
For example, \texttt{On} and \texttt{HandEmpty} are two predicates in the Ball-Ring environment, and $\abstractfn(\state_t)$ could be \{$\texttt{On(ball}, \texttt{floor)}, \texttt{HandEmpty(robot), \dots}$\}.
We additionally assume that each goal $\goal$ is associated with a set of ground predicates, e.g., $\{$\texttt{On(ball}, \texttt{table)}$\}$, which is a subset of $\abstractfn(\state_t)$ if and only if $\goal(\state_t) = 1$.

Next, for each parameterized skill, we assume access to a PDDL~\cite{fox2003pddl2} planning operator with predicate-based preconditions and effects.
For example, the operator for \texttt{Place(ball},~\texttt{table},~$\circ$\texttt{)} is:

\begin{Verbatim}[frame=single,resetmargins=true]
    Place(ball table)
    :precondition (and 
      (Holding ball)
      (Reachable table)) 
    :effect (and
      (On ball table)
      (HandEmpty)
      (not (Holding ball)))
\end{Verbatim}

Note the absence of continuous parameters.
The operator preconditions characterize the initiation condition of the skill.
For example, for the skill above, $\precondition(\state) = 1$ if $\{\texttt{Holding(ball)}, \texttt{Reachable(table)}\} \subseteq \abstractstate$ where $\abstractstate = \abstractfn(\state)$.
Similarly, the effects characterize the success condition of the skill.
Continuing the example, $\successfn(\state') = 1$ if $\{\texttt{On(ball}, \texttt{table}), \texttt{HandEmpty()}\} \subseteq \abstractstate'$ and $\{\texttt{On(ball}, \texttt{table}), \texttt{Holding(ball)}\} \not\in \abstractstate'$ where $\abstractstate' = \abstractfn(\state')$.
Previous work~\cite{silver2021learning,silver2023predicate,kumar2023learning} has learned operators and predicates; we manually specify them here.

Given an initial state $\state_0$ and a goal $\goal$, we construct a PDDL planning problem with initial state $\abstractfn(\state_0)$ and use a PDDL planner~\cite{helmert2006fast} to efficiently generate a sequence of planning operators (a skeleton) that chain together to reach the goal.
To incorporate skill competences, we associate a cost of $-\log{(\competence)}$ to a skill with competence $\competence$ and find a minimum-cost (maximum-likelihood) skeleton.
In experiments, we use LM-Cut (alias \texttt{seq-opt-lmcut} in Fast Downward) for minimum-cost planning.
We use a planning timeout of 10 seconds.
In experiments, this timeout was never triggered.

As explained in Section~\ref{subsec:planning}, once a skeleton is obtained, we greedily select continuous parameters using parameter policies and execute the resulting action.
Re-planning is triggered when the success condition (as defined by the operator effects) fails.
During evaluation, the robot continues planning and executing until a maximum number of actions $\horizoneval$ is reached (see Appendix~\ref{appendix:experiments}).

% To convert these operators into actions that can be executed, we need to determine continuous parameters for the corresponding parameterized skills.
% For example, the PDDL planner may suggest that we should execute \texttt{Place(ball},~\texttt{robot},~$\cdot$\texttt{)}, with a ``hole'' for the relative placement $\theta \in \mathbb{R}^2$ that we now need to fill in.
% We do this by greedily using our parameter policies (or parameter priors if on parameter policies have yet been learned), as explained in Section \ref{subsec:planning} and Algorithm \ref{alg:planning-to-solve-tasks}.

\subsection{Competence Models}
\label{appendix:comp-estim-details}

\begin{figure}[!h]
  \centering
    \noindent
    \includegraphics[width=0.95\columnwidth]{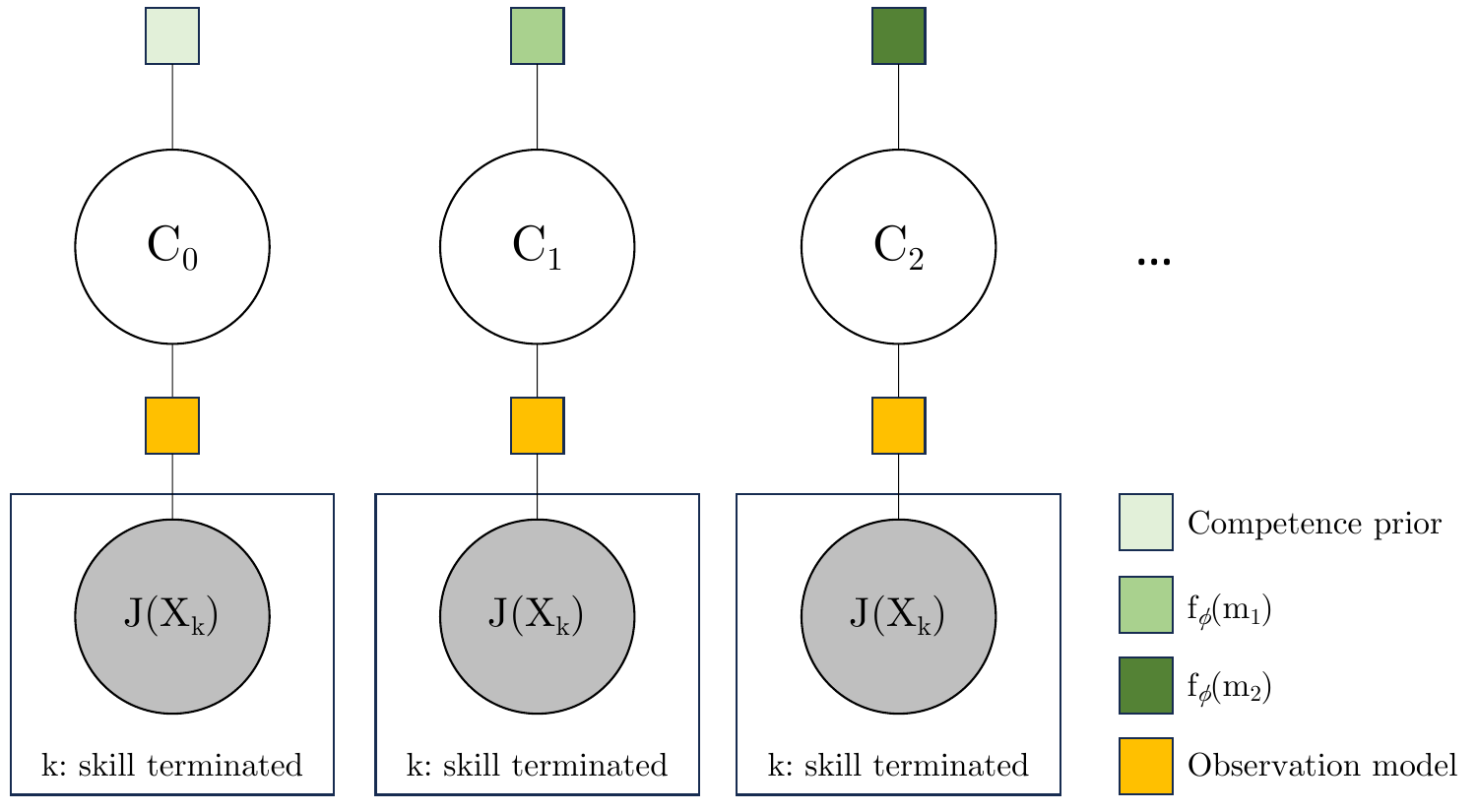}
    \caption{\textbf{Skill competence graphical model.}}
    \vspace{-0.5em}
  \label{fig:competence-model}
\end{figure}

We now provide details on the competence estimation and extrapolation methods described in Section~\ref{sec:selecting-skills-to-practice}.
Recall that each skill's competence is estimated and extrapolated independently.
In the main paper, we presented a graphical model relating the competence, success condition observations, and learning cycles of a skill; that model is summarized Figure~\ref{fig:competence-model}.

%%%%%%%%%%%%%%%%%%%%%%%%%%%%%%%%%%%%%%%%%%%%%%%%%%%%%%%%%%%%%%%%%%%%%%%%%%%%%%%%%%%%%%%%%%%%%%%%%%%%%%%
%%%%%%%%%%%%%%%%%%%%%%%%%%%%%%%%%%%%%%%%%%%%%%%%%%%%%%%%%%%%%%%%%%%%%%%%%%%%%%%%%%%%%%%%%%%%%%%%%%%%%%%

\noindent As described in the main text, the model components include:
\begin{itemize}
    \item The observation model $P(\successfn(\staterv_{i,k}) = 1 \mid \competencerv_i = \competence) = \competence$.
    \item The initial cycle prior $P_0(\competencerv_0)$; we use $\text{Beta}(10, 1)$.
    \item The other cycle priors $P_i(\competencerv_i) = f_\phi(m_i)$ where $m_i$ is the number of data collected for the skill through cycle $i$ and $f_\phi$ is a learned function that outputs Beta distributions.
\end{itemize}
We now discuss $f_\phi : \mathbb{Z}_{\ge 0} \to \text{Beta}(\alpha, \beta)$, the \emph{competence model}, in detail.
As mentioned, we considered two approaches for learning competence models: one based on expectation-maximization and another based on a simple sliding window.
After preliminary analysis, we opted to use the latter in our main experiments, but we describe both here for reference.

\emph{Approach 1: Expectation-Maximization.}
The model in Figure~\ref{fig:competence-model} is a latent variable model, so EM is a natural choice.
The E step is straightforward: since we have a collection of Beta-Bernoulli distributions, there is a closed-form solution to infer MAP competences given fixed competence models.
The M step---fitting the competence models given the most recent MAP competences---is more involved.

To start, we need to find a model class $\mathcal{F}$ with the property that any model in the class is non-decreasing in terms of the modes of the Beta distributions output by the model.
(Recall that this property is desirable because we do not want the robot to ever predict that practicing a skill will cause that skill to get worse; that skill would never be practiced.)
We use two ideas to satisfy this property.
First, rather than having our model output the Beta distribution parameters ($\alpha$ and $\beta$) directly, we output the mode and variance and then use those to derive the parameters.
Second, for predicting the mode, we pick a model class that is non-decreasing (and bounded between 0 and 1).
In our preliminary experiments, we used the exponential function $$f_\phi(m) = \phi_0 + (\phi_1 - \phi_0)(1 - \exp{(-\phi_2 m))}$$
where $\phi = [\phi_0, \phi_1, \phi_2]$, $0 \le \phi_0 \le 1, \phi_0 \le \phi_1 \le 1$, and $\phi_2 > 0$.
Note that $f_\phi(0) = \phi_0$, $\lim_{m \to \infty} f_\phi(m) = \phi_1$, and $\phi_2$ controls the rate of increase.

\begin{figure}[!ht]
  \centering
    \noindent
    \includegraphics[width=0.95\columnwidth]{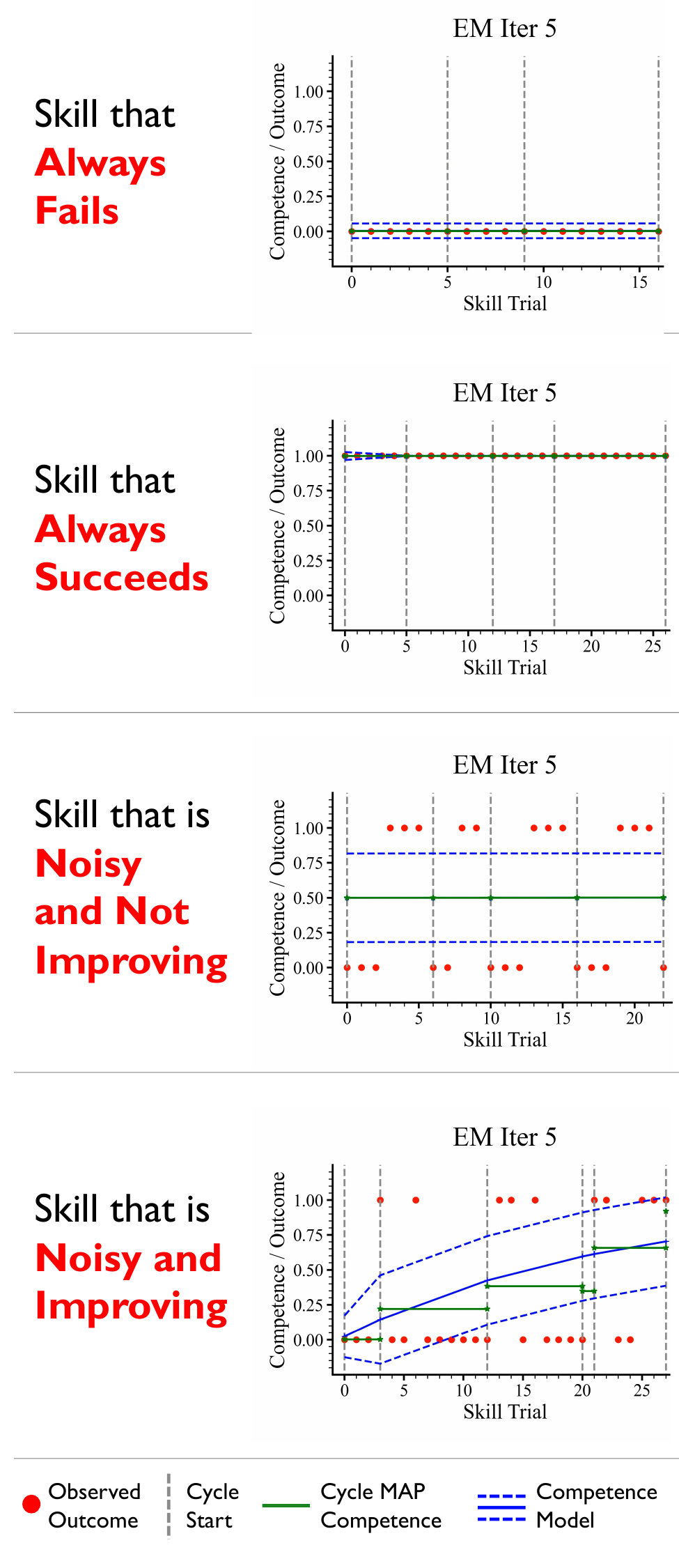}
    \caption{\textbf{Fitting competence models with EM.} For the competence models, solid lines are modes are dashed lines are variances.}
    \vspace{-1.5em}
  \label{fig:competence-model-em}
\end{figure}

To fit the model, we minimize a loss function $L(\phi) = \sum_i \ell_\phi(\hat{\competence}_i)$ where $\hat{\competence}_i$ is the MAP competence for cycle $i$ and $\ell_\phi : [0, 1] \to \mathbb{R}$ is the negative log likelihood under $f_\phi$.
Many techniques are possible to find $\phi^* = \argmin_\phi L(\phi)$, especially since $\phi$ is only three-dimensional.
In preliminary experiments, we used \texttt{scipy.optimize}.
The results of five iterations of EM on illustrative examples are shown in Figure~\ref{fig:competence-model-em}.
The performance is good in these examples, but EM is less stable in more realistic cases, which motivates the next method.

%%%%%%%%%%%%%%%%%%%%%%%%%%%%%%%%%%%%%%%%%%%%%%%%%%%%%%%%%%%%%%%%%%%%%%%%%%%%%%%%%%%%%%%%%%%%%%%%%%%%%%%
%%%%%%%%%%%%%%%%%%%%%%%%%%%%%%%%%%%%%%%%%%%%%%%%%%%%%%%%%%%%%%%%%%%%%%%%%%%%%%%%%%%%%%%%%%%%%%%%%%%%%%%

\emph{Approach 2: Simple Sliding Window.}
For the experiments in the main paper, we use a much simpler competence model that looks at the recent history of changes in competence and optimistically predicts that the best previous increase in competence between learning cycles will be repeated on subsequent cycles.
Concretely, $f_\phi(m+1)$ outputs a Beta with mode $\hat{\competence}_m + \max_{m - w \le i,j < m}(\hat{\competence}_j - \hat{\competence}_i)$, with clipping to enforce $f_\phi(m+1) \le 1$, where $w=2$ is the window size.
(The variance of the Beta distribution is not used during extrapolation.)
This approach is quite naive, especially considering that the number of data in each learning cycle is not directly taken into account.
Nonetheless, the approach worked well in experiments.

\subsection{Learning Parameter Policy Details}
\label{appendix:param-policy-details}
Recall from Section \ref{subsec:learning} that we learn an energy function $\energyfn_\parameterizedskill: \states \times \actions \to \mathbb{R}_{\ge 0}$ for each skill given a dataset of skill executions (including continuous parameters $\theta$) and success or failure of the $\successfn_\parameterizedskill$ check: $\dataset_\parameterizedskill = \{((\state_t, \action_t),\successfn_\parameterizedskill(\state_{t+1})) : \action_t \text{ uses skill } \parameterizedskill\}$.
We do this by training a two-layer Multi-Layer Perceptron (MLP) with hidden layers of size $32, 32$ with Binary Cross Entropy (BCE) loss.
We use the Adam \citep{kingma2014adam} optimizer with a learning rate of $10^{-3}$ for $10000$ iterations, or until $5000$ iterations have passed without any loss change.

As mentioned in the main text, we (1) share weights between skills with the same ``parent'' and (2) we construct a low-dimensional feature space to facilitate rapid learning.
To accomplish (1), we include object IDs in the low-level features and train one neural network per parent.
For example, for \texttt{Place}, we include features like $[0, 1]$ and $[0, 2]$ to distinguish \texttt{Place(ball},~\texttt{table},~$\circ$\texttt{)} from \texttt{Place(ball},~\texttt{floor},~$\circ$\texttt{)}), where $0$ is the ID for \texttt{ball}, etc.
For the rest of the features (2), we default to including the complete set of features for all objects that are included in the skill object parameters.
For certain skills that we know to be important for our experiments, we design the feature spaces more carefully:
\begin{itemize}
    \item In Ball-Ring, for \texttt{Place(ring},~\texttt{table},~$\circ$\texttt{)}, we include (1) the table size; (2) the x position of the rough patch; (3) the y position of the rough patch; (4) the size of the rough patch; (5) the x position of the table; (6) the y position of the table; and (7) the 2D continuous skill parameters themselves, which represent a relative placement on the table.
    \item In Cleanup Playroom, for all \texttt{Sweep} skills, we include the x and y positions of the object being swept, the x and y position of the bin, and the 1D continuous skill parameter, which represents a sweeping velocity.
    \item For all \texttt{Drop} skills in Cleanup Playroom, we use the 2D parameters alone, which represent a xy position for the gripper relative to the container before the gripper opens.
    \item For all \texttt{Pick} skills in the simulated version of Cleanup Playroom, we use the 2D continuous skill parameters alone, which represent a pixel in a canonical view of the object being picked. See Appendix~\ref{appendix:experiments} for further context.
\end{itemize}

\begin{figure*}[t]
  \centering
    \noindent
    \includegraphics[width=\textwidth]{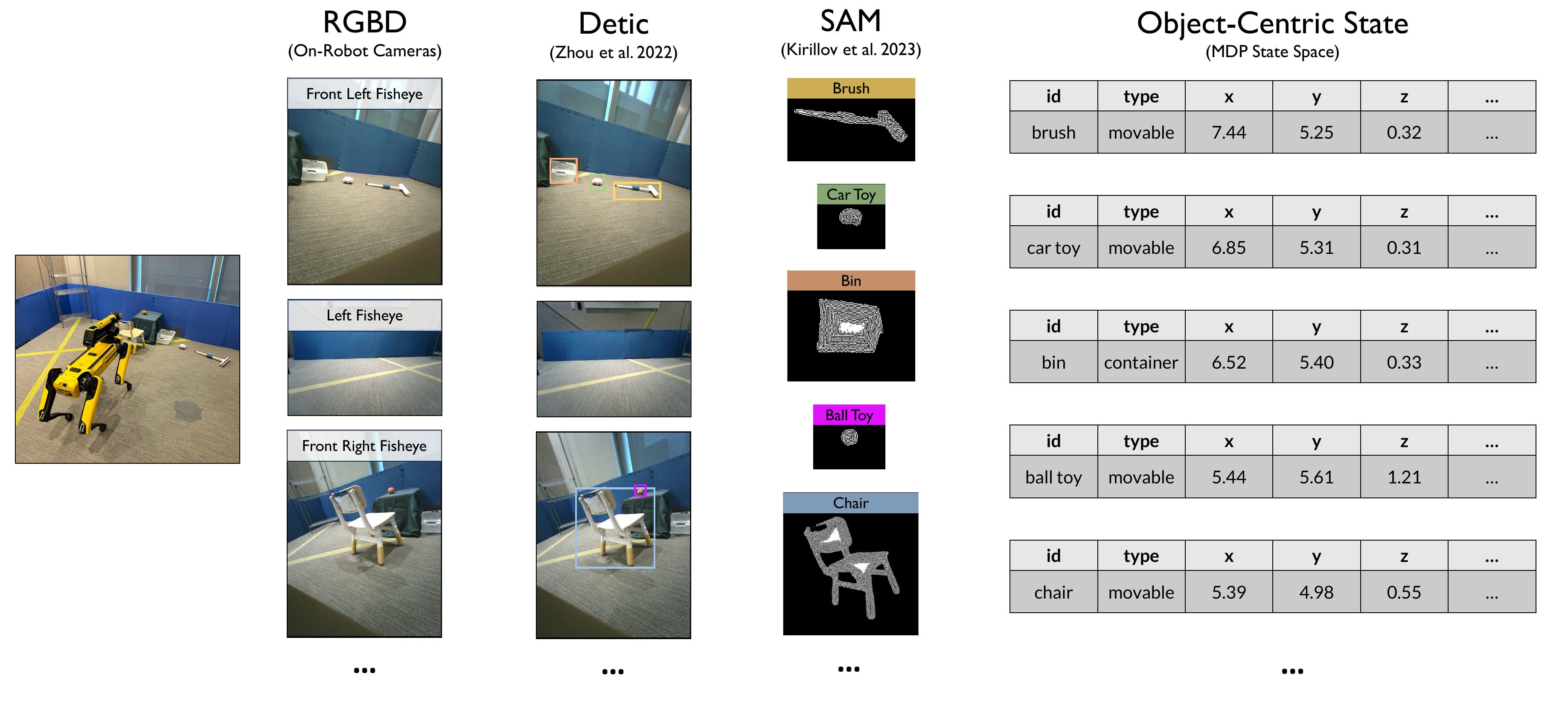}
    \caption{\textbf{Overview of perception pipeline.} We take RGBD images from Spot's cameras and then use Detic~\cite{zhou2022detecting} and SAM~\cite{kirillov2023segment} to construct an object-centric state. See text for details.}
  \label{fig:perception}
\end{figure*}

\subsection{Real Robot System Implementation Details}
\label{appendix:robot-details}

In this section, we provide system details for our real-robot experiments.
Recall that our model of the world (Section~\ref{subsec:world-modelling}) comprises object-centric states and parameterized actions.
To actualize this model, we need a \emph{perception system} that constructs object-centric states from sensors and \emph{skills} that can be executed on the robot.
All real-robot experiments use the Boston Dynamics Spot robot with an arm; see \url{https://bostondynamics.com/products/spot/} for specifications.

\subsubsection{Real-Robot Perception System}
Our perception system has three major components: (1) localization, (2) object detection; and (3) lost object search.

\emph{Localization.} To implement all our movement skills in a consistent manner that persists between runs, we assume the robot has access to a pre-defined map of its environment.
We construct this map by leveraging the Simultaneous Localization and Mapping (SLAM) stack that is part of the Boston Dynamics Spot SDK (see \url{https://dev.bostondynamics.com/docs/concepts/autonomy/graphnav_map_structure}).
This currently requires placing a number of fiducials around the environment.
Given a pre-defined map (which defines a coordinate system centered at the point where the map recording was begun), we implement a \texttt{localize()} method that gives us the current location of the robot within the map.
We leverage this functionality not only for movement (discussed in the skills subsection below), but also to compute the positions of objects we see in the world, since they are detected relative to the robot's cameras (discussed in the object detection subsection below).
The position of the robot itself is also added to the object-centric state.
Within the map, we also define the boundaries of all ``allowed'' regions where the robot can navigate to (disallowed regions include obstacles not captured during mapping such as clear glass walls, etc.). When navigating, we only allow the robot to move to a position that is within the convex hull of points that define the periphery of an allowed region.
Additionally, if we detect an object to not be inside an allowed region, we automatically consider that detection invalid and throw it away.

\emph{Object Detection.}
The Spot robot collects RGBD images from six perspectives: one in the hand, two on the front, one on each side, and one on the back.
We collect all six images at each time step and then run object detection in each image.

To detect objects in an RGBD image, we use a combination of Detic~\cite{zhou2022detecting} and Segment Anything (SAM)~\cite{kirillov2023segment}.
See Figure~\ref{fig:perception} for a summary.
Detic uses CLIP embeddings to identify object bounding boxes in RGB images given natural language class names (``prompts'').
We experimented with a number of objects and prompts to find a combination that would work reliably in our setting, and even so, significant noise remains.
See Table~\ref{tab:prompt-hacks} for the final set of prompts used.
For each bounding box returned by Detic, we run SAM inside the bounding box to get a mask for the object.
The center of the bounding box gives the $xy$ position of the object in the camera frame.
We then compute the median depth value in the mask to get a $z$ value, and transform the full $xyz$ position into the world frame using the known camera intrinsics and robot pose from localization.
The rotation of the object is not detected.
Known object features (such as object size, whether it is movable, etc.) are added to the $xyz$ position features.
Static objects (e.g., tables) are added to the state automatically, rather than visually detected, for simplicity.
As a method for object detection, this overall approach has a number of limitations (see ``failure modes'' below).
However, one advantage is that it is fast---about 0.25 seconds overall---which is important since we are running it at every time step between skill executions.

When the same object is detected in multiple cameras at the same time step, we use the detection with the highest confidence score returned by Detic.
Object detections are aggregated over time: whenever an object is detected, the previous detection is overwritten.
More sophisticated state estimation strategies are possible.
To initialize object detections at the very beginning of online learning, the robot navigates to a fixed home pose, raises its arm to get a top-down view, and rotates in place, collecting images and detecting objects until all known objects have been seen.

\emph{Lost Object Search.}
We assume that all objects can be found in the initial scene by the simple rotate-in-place procedure described above, and that objects are not removed from the scene while the robot is running.
Even so, objects can become lost, and it is important for both learning and planning that the robot can find them.
A principled approach for object search under partial observability is outside the scope of this paper.
Instead, we use the following domain-specific logic.
First, when a pick skill is executed, if the robot's gripper is subsequently open beyond a threshold value, then we assume that the target object was successfully grasped and we update a corresponding \emph{held} feature in the object-centric state.
Then, if a place or drop skill is executed, and if the held object can be subsequently seen, the held feature is updated accordingly.
If a pick skill is executed and the gripper is subsequently closed, or if a place/drop skill is executed and the object is not subsequently seen, the object is declared \emph{lost}.
When an object is lost, the robot executes a special \emph{find objects} procedure.
This procedure starts by executing a series of fixed move-and-look actions and then begins to randomly sample move-and-look actions until the lost object is seen.
In rare cases where more than 10 move-and-look actions are executed and the object is still not found, we manually take control of the robot and point it to look at the lost object.
The \emph{find objects} procedure is executed externally from the approach; the transitions collected are not used for learning.

\begin{table}[t]
\resizebox{\columnwidth}{!}{%
\begin{tabular}{@{}ccc@{}}
\toprule
\textbf{Environment} & \textbf{Object} & \textbf{Prompt} \\ \midrule

Ball-Ring & ball &
  \begin{tabular}{@{}c@{}}
  small white ball / \\
  ping-pong ball / \\
  snowball / \\
  cotton ball / \\
  white button \\
  \end{tabular}

\vspace{0.25em} \\ \hline \\[-0.25em]

Ball-Ring & ring & \begin{tabular}{@{}c@{}}
  yellow hoop toy / \\
  yellow donut\\
  \end{tabular}

\vspace{0.5em} \\ \hline \\[-0.25em]

Cleanup Playroom & bin & \begin{tabular}{@{}c@{}}
  white plastic container with black handles / \\
  white plastic tray with black handles / \\
  white plastic bowl / \\
  white storage bin with black handles \\
  \end{tabular} 

\vspace{0.5em} \\ \hline \\[-0.25em]

Cleanup Playroom & ball toy & \begin{tabular}{@{}c@{}}
  small orange basketball / \\
  small orange \\
  \end{tabular} 

\vspace{0.5em} \\ \hline \\[-0.25em]

Cleanup Playroom & car toy & \begin{tabular}{@{}c@{}}
  small white ambulance toy / \\
  car\_(automobile) toy / \\
  egg \\
  \end{tabular} 

\vspace{0.5em} \\ \hline \\[-0.25em]

Cleanup Playroom & platform & \begin{tabular}{@{}c@{}}
  black coffee table / \\
  bench
  \end{tabular} 

\vspace{0.5em} \\ \hline \\[-0.25em]

Cleanup Playroom & chair & chair

\vspace{0.5em} \\ \hline \\[-0.25em]

Cleanup Playroom & brush & \begin{tabular}{@{}c@{}}
  scrubbing brush / \\
  hammer / \\
  mop / \\
  giant white toothbrush \\
  \end{tabular} 

\\ \bottomrule
\end{tabular}%
}
\caption{\textbf{Detic prompts.} The backslashes are included in the prompt; each row entry represents a single prompt.}
\label{tab:prompt-hacks}
\vspace{-1.5em}
\end{table}

\subsubsection{Real-Robot Skills}
We implement skills for the Spot robot on top of the Boston Dynamics SDK (\url{https://dev.bostondynamics.com/}).
All skills are listed in Appendix~\ref{appendix:experiments} and implementations are given in the code accompanying the paper.
(see \url{http://ees.csail.mit.edu}).
We discuss two types of skills here---\emph{move} and \emph{pick}---and refer to the code for others.

\emph{Move skills.} The Spot SDK provides functionality for moving the robot base to a relative SE(2) pose.
Collisions are anticipated and a certain amount of local navigation around obstacles is handled automatically.
Because of this, and because our environments are relatively free of obstacles, we do not require full-fledged motion planning.
However, we do need to sample collision-free target positions to implement the parameterized move skills.
For example, the parameter prior for \texttt{MoveTo} skills samples a distance and an angle relative to a target object; only collision-free poses should be sampled.
We perform conservative collision checking in SE(2) using the known robot and object dimensions.

\textit{Pick skills.} The Spot SDK provides functionality for grasping at a pixel in the hand camera image. Rotation constraints on the gripper can also be enforced.
By default, our \texttt{Pick} skills select a random pixel in the target object mask returned by Detic/SAM, with no rotation constraints enforced.
However, for certain objects, we implement specific grasp pixel selection logic.
In principle, such logic could be learned, but doing so would require learning over images, which we do not consider in this work.
Note that we \emph{do} learn nontrivial parameter policies for grasping in the simulation experiments; see Appendix~\ref{appendix:experiments}.
As an example of object-specific grasp selection, for the brush in Cleanup Playroom, we do the following:
(1) Detect the largest connected component of blue pixels in the image (i.e., the center of the brush handle);
(2) Choose the center pixel of that connected component for grasping;
(3) Find the head of the brush with respect to the center of the pixel and set the rotation constraint so that the head is to the right of the gripper.
The logic for other objects is typically less involved; see the accompanying code (\url{http://ees.csail.mit.edu}).

\subsubsection{Real-Robot Limitations and Failure Modes}
Computer vision, robot skill policies, and robot hardware are all continuing to improve, and the methods proposed in this paper will continue to be applicable as they do.
However, the stack we implemented has several limitations.
In this section, we mention a few of those limitations and note how we worked around them for the purpose of running our experiments.

\emph{Selecting objects for Cleanup Playroom.}
We selected the ball and car toys used in the Cleanup Playroom environment after considering multiple constraints.
The objects need to be small enough for the robot to grasp, large enough that a successful grasp can be distinguished from a fully-closed gripper, light enough for the robot to lift and sweep, heavy enough that they won't always bounce out of the bin when swept, and visually distinct enough that Detic can reliably find them (on the table, on the floor, in the bin, etc.).

\emph{Selecting (short) tables and bins.}
The tables and bins used in both real-robot environments are notably low to the ground.
This is because the robot needs to see objects on top of the tables and inside the bins.
In earlier versions of our environments, we included ``look from above'' actions at the end of certain skills, like placing objects on tables, by moving the robot's hand high up and looking down.
But this is not possible in the Cleanup Playroom environment, where the robot is holding the brush after sweeping into the bin.
Another workaround would be to place cameras or other sensors in the external environment, but we were committed to working with Spot's on-board sensors alone.

\emph{3D printed parts and adding wheels to drag objects.}
The blue component on the brush in Cleanup Playroom is a part that we 3D printed and added on.
Without that part, the robot was not able to consistently grasp the brush from the floor and prepare to sweep without the brush slipping in-hand.
The grey `blocker' pieces at the bottom left of the table in our Ball-Ring environment are 3D printed parts intended to prevent the ring from sliding off the left side of the table despite being correctly placed there.
Additionally, the platform and chair in the Cleanup Playroom environment could not be reliably dragged without attaching specific ball-bearing wheels to their legs. These were specifically selected and attached to the objects with 3D printed parts.

\emph{Object detection failures.}
With Detic, we regularly encounter false positives (objects detected where they shouldn't be) and false negatives (objects not detected at all).
We deal with this by prompt hacking (Table~\ref{tab:prompt-hacks}) and discarding detections with confidences below a threshold (0.4).
Even when objects are detected correctly, there is another issue stemming from the fact that we do not assume known object models and we are not doing true pose detection.
Since we use bounding box centers to define the $xy$ position of the object in the camera frame, and since our views of the objects are constantly changing, the object's reference frame is also constantly changing.
Surprisingly, this is often not an issue; the skills we use in this work are largely robust to these variations.
But occasional failures can occur.
For example, suppose that the robot sees only the top of an object and assigns a $z$ position in the world frame that has the object floating above the floor.
The classifier for \texttt{On(obj,} \texttt{floor)} would subsequently misfire, leading to initiation condition failures in skills like \texttt{Pick(obj,} \texttt{floor)}.
In future work, we hope to integrate full-fledged pose detection into our system, but note the need for low latency and high accuracy.

\subsection{Approach Details}
\label{appendix:approach-details}

In this section, we give implementation details for each of the approaches used in the paper.
The following details are shared for all approaches.
We alternate between task time, free time, and learning.
During task time, the approaches pursue the given task goal until it is either achieved, or until $\horizoneval$ steps have been taken.
These task horizons are $\horizoneval=27$ (the number of grid cells plus 2) for Light-Switch, 8 for Ball-Ring (Simulated), 10 for Cleanup Playroom (Simulated), 15 for Ball-Ring (Real), and 12 for Cleanup Playroom (Real).
Free time then lasts until a maximum number of steps is reached: 150 for Light Switch, 100 for Ball-Ring (Simulated), 125 for Cleanup Playroom (Simulated), 20 for Ball-Ring (Real), and 50 for Cleanup Playroom (Real).
For the epsilon-greedy policy in skill practicing, we use $\epsilon=0.5$.
Learning details are given in the main paper.

\subsubsection{Planning-to-Practice Approaches}
There are five approaches that plan to practice: EES (Ours), Fail Focus, Competence Gradient, Skill Diversity, and Task-Relevant.
These approaches are identical except in their criteria for choosing a skill to practice.
EES and Task-Relevant call the planner internally; EES uses the planner for the ``situate'' step, and Task-Relevant uses the planner to determine which skills are relevant to previously seen tasks.
Re-invoking the planner on all previously seen tasks at each practice decision can be slow, so we use two optimizations.
First, we use only the 10 most recently seen tasks.
Second, we cache the last plan output for each task and rerun the planner only once out of every 100 calls.
Note that plans cannot be cached for all time because the changing skill competences may change the maximum-likelihood plans.
More sophisticated caching strategies that use the competence changes themselves are possible.

\subsubsection{MAPLE-Q}
\label{appendix:mapleq}
MAPLE \citep{nasiriany2022maple} requires three trained networks: (1) a \textit{task policy} network, (2) a \textit{parameter policy} network, and (3) a Q-network ($Q_{\sigma}(x, u, \theta)$).
Here, $x$ is an input state, $a$ is a parameterized skill, and $\theta$ is a set of continuous parameters to be input to a ground skill.
The task network is intended to select a skill $u$ and the parameter policy network is intended to select a continuous parameter vector $\theta$ conditioned on the ground skill $u$ and the state $x$ (similar to our skill parameter policies). Together, these networks serve as an actor that outputs a ground parameterized skill that can be executed in the environment.
The Q-network $Q_{\sigma}(x, u, \theta)$ serves as a critic that outputs Q-values given a state and skill.

Given our setting and main approach (EES), there are two significant reasons why comparing against MAPLE directly is unfair.
Firstly, our approach can leverage symbolic operators (specifically their preconditions) to discern states in which particular skills are applicable, whereas the task policy in MAPLE must learn this.
And secondly, our approach has access to parameter priors for each skill that can be used to produce policy parameters $\theta$, whereas the parameter policy network in MAPLE does not. 

We seek to remedy these by giving MAPLE access to our symbolic operators and parameter priors, and doing away with the task and policy networks.
Specifically, given a state $x$, we use our operators to determine which skills can be executed from this state. For each of these skills, we sample $n_{\text{s}}$ number of continuous parameter vectors from our parameter prior.
We then pass $n_{\text{s}}$ tuples of $(x, u, \theta)$ per applicable skill through the Q-network and choose the maximum. Intuitively, the Q-network must not only implicitly learn good parameter policies, but also how to sequence together skills given a goal (i.e., learning how to plan).

More concretely, we train a Q-network that takes as input a vector consisting of $4$ smaller vectors concatenated together: (1) a continuous vector of the features of all objects in the current state (i.e., state $x$), (2) a one-hot vector corresponding to the skill to be invoked from the current state, (3) a vector of the continuous parameters $\theta$ to be passed to the skill (potentially padded with zeros), (4) a one-hot vector corresponding to the current goal being solved (for environments with multiple possible goals). Given this input, the Q-network predicts a Q-value. Similar to our other approaches, we train the network with a batch size of $64$ using the Adam optimizer for $10000$ iterations, with early stopping after $5000$ iterations of no change in the loss. We use epsilon-greedy exploration during free-time, with epsilon set to $0.5$, and at test time, we sample $100$ $\theta$ vectors from the parameter prior of each skill applicable in the current state and run the skill with the maximum Q-value from these.

\subsection{Environment Details}
\label{appendix:experiments}
In this section, we detail the environments used in our experiments.
Note that all skills listed below have discrete object parameters (indicated by the `\texttt{?}') as well as continuous parameters $\theta$ (shown within []). Thus, the total number of skills is much larger than shown here, since these skills need to be ground with the various objects in the environment.
For specific implementation details beyond what is presented here, please see the accompanying code (\url{http://ees.csail.mit.edu}).

\begin{tightlist}
\item \textit{Light Switch (Simulated)}: The main challenge in this environment is for the robot to specialize its parameter prior for the \texttt{ToggleLight} skill.
    \begin{tightlist}
        \item Predicates: \texttt{RobotInCell(?robot, ?cell)}, \texttt{LightInCell(?light, ?cell)}, \texttt{LightOn(?light)}, \texttt{Adjacent(?cell1, ?cell2)}
        \item Skills: 
        \begin{tightlist}
            \item \texttt{MoveTo(?robot, ?cell1, ?cell2)}: Moves the robot between \texttt{?cell1} and \texttt{?cell2} provided the robot is currently in \texttt{?cell1} and \texttt{?cell2} is adjacent to it.
            \item \texttt{ToggleLight(?robot, ?light, [dlight])}: Spins the light dial if the robot is currently in the same cell as the light. If the `level' feature of the light plus the \texttt{dlight} continuous parameter value yields the \texttt{target} value for the light (which is a feature of the light), then the light will turn on. The parameter prior for \texttt{dlight} is simply a uniform distribution over $[0, 2\pi]$.
            \item \texttt{JumpToLight(?robot, ?cell1, ?cell2, ?cell3, ?light)}: Tries to have the agent `jump' directly from \texttt{?cell1} to \texttt{?cell3} given cells 1, 2, and 3 are adjacent. However, this skill is impossible and never achieves its purported effect.
        \end{tightlist}
        \item Goal(s): Achieve \texttt{LightOn(?light)} given the robot starts in the first cell.
    \end{tightlist}
\item \textit{Ball-Ring (Simulated)}: Note that there is only one robot, one ball and one ring, but 5 different tables, some of which are slanted and others which are not.
The main challenges in this environment are for the robot to learn that the competence of the \texttt{Place(ball, table)} skill is very low for a slanted table and cannot be improved (thus requiring it to switch its strategy for accomplishing the goal), and for it to specialize the parameter prior for the \texttt{Place(ring, table)} skill to place the ring on the high-friction part of the table so it doesn't slide down.
    \begin{tightlist}
        \item Predicates: \texttt{On(?obj, ?surface)}, \texttt{Reachable(?robot, ?obj)}, \texttt{Inside(?obj, ?container)} 
        \item Skills: 
        \begin{tightlist}
            \item \texttt{Pick(?robot, ?obj, [x, y])}: Picks up an object if the robot is reachable to it and the robot's hand is currently empty. The pick will only succeed if the [x, y] params fall somewhere on the object's surface. The parameter prior is designed to be perfect, leading to 100\% success at this action.
            \item \texttt{PlaceOnTop(?robot, ?obj, ?surface, [x, y])}: If the robot is holding \texttt{?obj} and reachable to \texttt{?surface}, this skill will place the ball at the [x, y] parameters indicated. Note that it is not possible to make the ball stay on any table (it will always bounce/roll off). The parameter prior is a uniform distribution over all locations on \texttt{?surface}.
            \item \texttt{PlaceInside(?robot, ?obj, ?container, [x, y])}: Similar to the above \texttt{PlaceOnTop} skill, but instead attempts to place \texttt{?obj} inside \texttt{?container}. The parameter prior is uniform over the inside surface of \texttt{?container}, which yields success 100\% of the time when called from an applicable state.
            \item \texttt{NavigateTo(?robot, ?obj, [x, y])}: Moves the robot to the [x, y] parameters indicated, unless they are in collision with some object. The parameter prior is a uniform circle around \texttt{?obj} such that the radius of the circle is the maximum distance at which the robot will be reachable to the object.
        \end{tightlist}
        \item Goal(s): Achieve \texttt{On(ball, table1)} for a particular \texttt{table1} that happens to be slanted, which means the only way to succeed is to place the ring on the table, and then place the ball in the ring. The ball always starts out atop one of the tables, while the ring is on the floor. Table positions are randomized around the room.
    \end{tightlist}
\item \textit{Real-World Domains and Cleanup Playroom (Simulated)}: These environments all share a common set of predicates and skills. We list these first, followed by the skills and goals specific to each particular environment.
    \begin{tightlist}
        \item Predicates: \texttt{NotEqual(?obj0, ?obj1)}, \texttt{OnTop(?obj0, ?obj1)}, \texttt{Above(?obj0, ?obj1)}, \texttt{Inside(?obj0, ?obj1)}, \texttt{FitsInsideXY(?obj0, ?obj1)}, \texttt{HandEmpty(?robot)}, \texttt{Holding(?robot, ?obj0)}, \texttt{InHandView(?robot, ?obj0)}, \texttt{Reachable(?robot, ?obj0)}, \texttt{Blocking(?obj0, ?obj1)},  \texttt{IsPlaceable(?obj0)}, \texttt{IsSweeper(?obj0)}, \texttt{HasFlatTopSurface(?obj0)}, \texttt{PlatformInFrontOfSurface(?platform, ?surface)}, \texttt{SurfaceTooHigh(?surface)}, \texttt{SurfaceNotTooHigh(?surface)}, \texttt{RobotOnPlatform(?robot)}
        \item Skills:
        \begin{tightlist}
            \item \texttt{MoveToReachObject(?robot, ?obj0, [dist, angle])}: Moves the robot to a position that is \texttt{dist} meters away and \texttt{angle} radians from the center location of \texttt{?obj0} and is oriented such that it is facing the center of the object. The parameter prior is a uniform distribution over distances between 0.1 and 0.8 meters, and angles between 0 and $2\pi$ radians.
            \item \texttt{MoveToViewObject(?robot, ?obj0, ?surface, [dist, angle])}: Same as the \texttt{MoveToReachObject}, except that it also moves the robot's hand such that it is gazing directly at \texttt{?obj0}. The parameter prior is also set to be uniform in distances between 1.1 and 1.3m (further away than \texttt{MoveToReachObject}) so the hand camera gets a clear view of the whole object.
            \item \texttt{PickObjectFromTop(?robot, ?obj0, ?surface, [px, py, qx, qy, qz, qw])}: If \texttt{?obj0} is not too-high up for the robot to reach, this skill tries to pick it up by grasping at the pixel location [px, py] in the image taken from the hand camera prior to grasping and with the arm oriented according to the quaternion \texttt{[qx, qy, qz, qw]}. For most objects, the parameter prior is uniform over all pixels in the object mask, and there is no constraint places on the quaternion (allowing the skill to grasp the object in whatever orientation it chooses). For some objects, such as the chair, the parameter prior is a much narrower distribution (e.g. for the chair, \texttt{px, py} are constrained to be a point at the top of the chair back, and the quaternion is constrained to do a top down grasp).
            \item \texttt{PlaceObject(?robot, ?obj0, ?surface, [dx, dy, dz])}: If the robot is currently holding an object, moves the hand to \texttt{[dx, dy, dz]} from the object's top-most center point and opens the gripper. This generally leads to a very high success rate for placing objects.
            \item \texttt{PickAndDumpContainer(?robot, ?container, ?surface, ?obj, [px, py, qx, qy, qz, qw])}: Similar to the \texttt{PickObjectFromTop}, but picks up \texttt{?container} that contains \texttt{?obj0}, and then tries to dump \texttt{?obj0} so it falls onto the floor before putting \texttt{?container} back where it was picked from.
        \end{tightlist}
    \end{tightlist}

%%%%%%%%%%%%%%%%%%%%%%%%%%%%%%%%%%%%%%%%%%%%%%%%%%%%%%%%%%%%%%%%%%%%%%%%%%%%%%%%%%%%%%%%%%%%%%%%%%%%%%%
\begin{figure}[t]
  \centering
    \noindent
    \includegraphics[width=0.8\columnwidth]{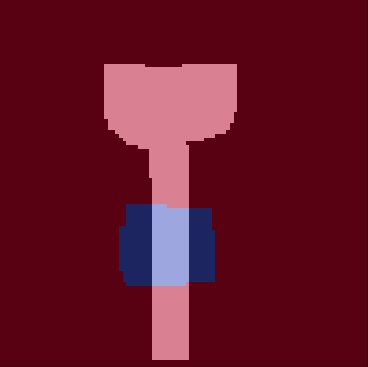}
    \caption{\textbf{Visualization of grasping constraints for Simulated Cleanup Playroom environment}. The top-down profile of the brush object is shown in pink, while the allowed region for grasping is shown in blue. If the robot attempts to grasp the object outside the allowed region, the grasp will fail.}
    \vspace{-1.5em}
  \label{fig:brush-grasp}
\end{figure}
%%%%%%%%%%%%%%%%%%%%%%%%%%%%%%%%%%%%%%%%%%%%%%%%%%%%%%%%%%%%%%%%%%%%%%%%%%%%%%%%%%%%%%%%%%%%%%%%%%%%%%%

\item \textit{Real-World Ball-Ring}: The main challenges of this environment are identical to the simulated variant. Given that this is implemented with real perception and control, there is also non-trivial noise to be dealt with. Note that all skills used here are listed in the above entry.
    \begin{tightlist}
        \item Goal: Achieve \texttt{BallOnTable(?ball, table1)} where \texttt{table1} is the slanted table in the room. Unlike the simulated version, there is only one table, and it is slanted, with about 40\% of the left side covered with a high-friction material such that the ring will stay there if placed, and the remainder left smooth so that the ring will slide down immediately.
    \end{tightlist}

%%%%%%%%%%%%%%%%%%%%%%%%%%%%%%%%%%%%%%%%%%%%%%%%%%%%%%%%%%%%%%%%%%%%%%%%%%%%%%%%%%%%%%%%%%%%%%%%%%%%%%%
\begin{figure*}[t]
  \centering
    \noindent
    \includegraphics[width=0.98\textwidth]{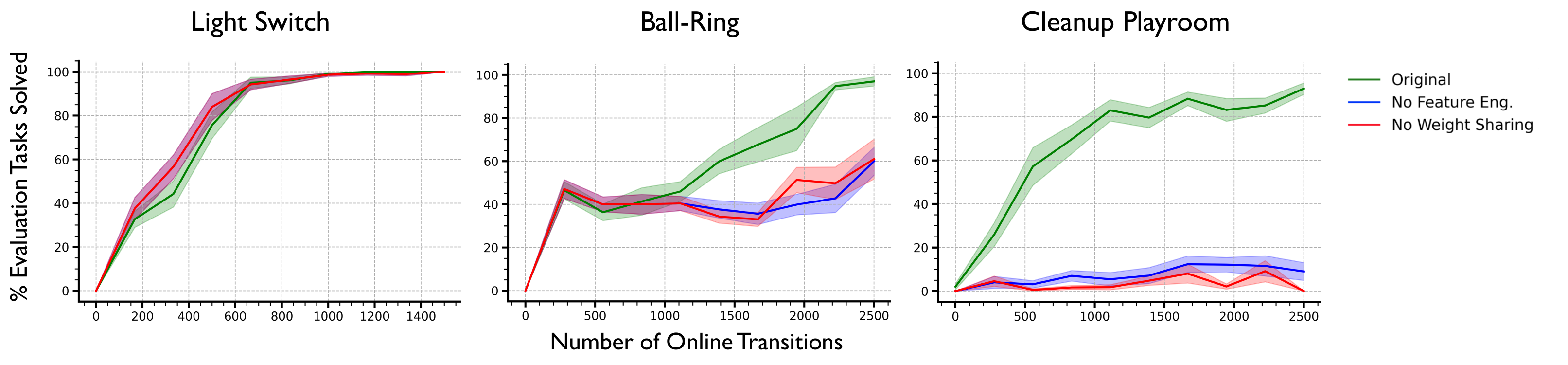}
    \caption{\textbf{Ablation Experiment Results.} Percentage of evaluation tasks solved vs. number of online transitions collected for our EES approach in simulated environments under two ablations. The ``No Feature Eng.'' baseline ablates all implemented feature engineering (note that there was no feature engineering for the original Light Switch environment results). The ``No Weight Sharing'' baseline ablates neural network weight sharing. Solid lines represent means and shading represents standard error across 10 seeds.}
  \label{fig:additional-sim-results}
\end{figure*}
%%%%%%%%%%%%%%%%%%%%%%%%%%%%%%%%%%%%%%%%%%%%%%%%%%%%%%%%%%%%%%%%%%%%%%%%%%%%%%%%%%%%%%%%%%%%%%%%%%%%%%%

\item \textit{Cleanup Playroom (Simulated)}: The main challenges in this environment are specializing the parameter prior to correctly learn to grasp the brush (and other objects) in their allowed regions (see Figure \ref{fig:brush-grasp}), and learning to specialize the parameter prior for the velocity to use to sweep objects successfully from various positions atop a table into a bucket.
    \begin{tightlist}
        \item Skills (in addition to those in the entry on `Real-World Domains and Cleanup Playroom (Simulated)`):
        \begin{tightlist}
            \item \texttt{DragToUnblockObject(?robot, ?blocker, ?blocked, [dx, dy, dyaw])}: if the robot is grasping \texttt{?blocked} and \texttt{?blocked} is also being blocked by \texttt{?blocked}, then this skill will move \texttt{?blocker} to unblock \texttt{?blocked}. This is accomplished by moving the robot by \texttt{dx, dy} in the x and y directions respectively before rotating by \texttt{dyaw}. The parameter prior is set to simply always rotate by a specific amount, since this is often enough to unblock objects in our environment.
            \item \texttt{DragToBlockObject(?robot, ?blocker, ?blocked, [dx, dy, dyaw])}: Does the exact opposite of the \texttt{DragToUnblockObject} skill provided the robot starts out having already grasped \texttt{?blocker}.
            \item \texttt{SweepIntoContainer(?robot, ?sweeper, ?obj, ?surface, ?container, [velocity])}: If the robot is holding \texttt{?sweeper}, is reachable to \texttt{?surface)}, \texttt{?container} is ready for sweeping, and \texttt{?obj} is atop \texttt{?surface}, then uses \texttt{?sweeper} to try to push \texttt{?target1} and \texttt{?target2} into \texttt{?container}. The sweeping motion is performed with velocity corresponding to the \texttt{velocity} parameter. The parameter prior is a uniform distribution over a range of velocities the robot arm is capable of moving at.
        \end{tightlist}
        \item Goal(s): Achieve \texttt{ObjectInsideContainer} for two different objects that begin atop the sole table in the environment. We also experimented with only having the goal mention one object, which drastically changes the maximum likelihood task plan. 
    \end{tightlist}
    \item \textit{Real-World Cleanup Playroom}: The main difference between this environment and its simulated counterpart is that the parameter priors already provide a very high success rate for grasping the brush and other objects (since learning a grasp sampler in the real world is prohibitively sample inefficient). The main challenge is thus learning to specialize the parameter prior for sweeping real objects from the table into the bucket. Given that this is implemented with real perception and control, there is also non-trivial noise to be dealt with.
    \begin{tightlist}
        \item Skills (in addition to those from the simulated variant of this environment):
        \begin{tightlist}
            \item \texttt{DragPlatformInFrontOfSurface(?robot, ?platform, ?surface)}: If the robot is initially grasping \texttt{?platform}, drags it such that it is positioned in front of \texttt{?surface} so that the robot can stand atop it to pick up an object atop \texttt{?surface} (which is too high for the robot to ordinarily pick objects from).
        \end{tightlist}
        \item Goal: Achieve \texttt{ObjectInsideContainer} for two different objects that begin atop the sole table in the environment.
    \end{tightlist}
\end{tightlist}

\subsection{Additional Experiments}
\label{appendix:additional-experiments}
In order to investigate the impact of feature engineering (Appendix Section \ref{appendix:param-policy-details}) and weight sharing among parameter policy networks (Section \ref{subsec:learning}) on our approach, we ran additional experiments ablating each of these choices respectively. The results are displayed in Figure \ref{fig:additional-sim-results}.

As can be seen from Figure \ref{fig:additional-sim-results}, weight sharing has a negligible impact on performance in the simple Light-Switch environment. In the more-complex Ball-Ring environment, both weight sharing and feature engineering have a significant impact. Our approach reaches only about 50\% success rate without each of these as compared to the 100\% success rate it achieves in the same number of steps with both of these implemented. Finally, in the challenging `Cleanup Playroom' environment, we find that both weight sharing and feature engineering are critical to our approach's performance. Without either of these, our approach is not able to meaningfully improve success rate on test tasks even after 2000 steps. These findings suggest that feature engineering and weight sharing become increasingly critical to good performance as the complexity of the task under consideration increases.

\end{document}